\documentclass[10pt,twocolumn,letterpaper]{article}

\usepackage{cvpr}
\usepackage{times}
\usepackage{epsfig}
\usepackage{graphicx}
\usepackage{amsmath}
\usepackage{amssymb}
\usepackage{amsthm}
\usepackage{csquotes}
\usepackage{mathtools}
\DeclarePairedDelimiter{\norm}\lVert\rVert
\usepackage{animate}
\usepackage{floatrow}
\usepackage{authblk}
\newcommand*\samethanks[1][\value{footnote}]{\footnotemark[#1]}

\usepackage[breaklinks=true,bookmarks=false]{hyperref}

\cvprfinalcopy 

\def\cvprPaperID{****} 

\begin{document}

\title{Learning Linear Transformations for Fast Arbitrary Style Transfer}

\author[1]{Xueting Li\thanks{Contributed equally.}}
\author[2]{Sifei Liu\samethanks}
\author[2]{Jan Kautz}
\author[1]{Ming-Hsuan Yang}

\affil[1]{University of California, Merced}
\affil[2]{NVIDIA}

\maketitle

\begin{figure*}[t]

	\centering

	\begin{tabular}{c@{\hspace{0.005\linewidth}}c@{\hspace{0.005\linewidth}}c@{\hspace{0.005\linewidth}}c@{\hspace{0.005\linewidth}}c}
    	\includegraphics[height = .173\linewidth, width = .260\linewidth]{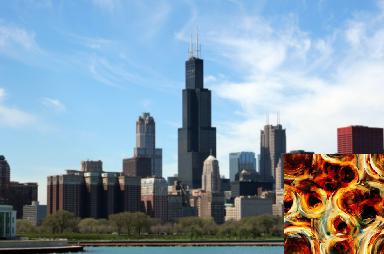} &
		\includegraphics[height = .173\linewidth, width = .260\linewidth]{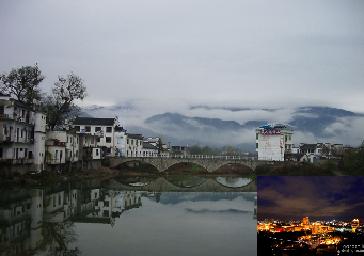} &
		\animategraphics[autoplay,loop,height = .173\linewidth, width = .173\linewidth]{5}{images/teaser/content/}{1}{39}&
		\includegraphics[height = .173\linewidth, width = .260\linewidth]{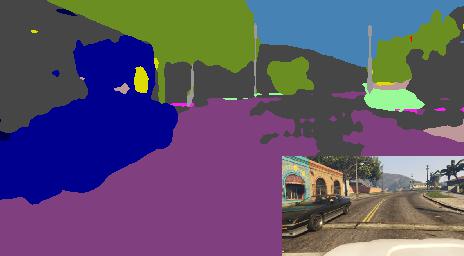} &\\
		
		\includegraphics[height = .173\linewidth, width = .260\linewidth]{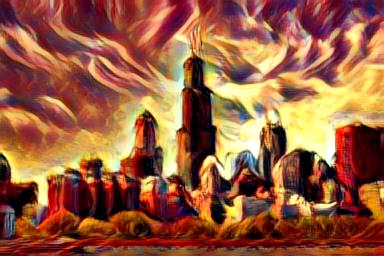} &
		\includegraphics[height = .173\linewidth, width = .260\linewidth]{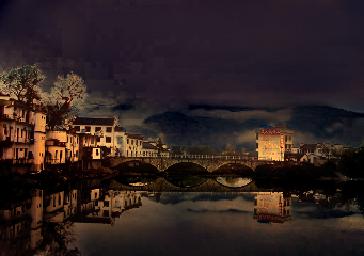} &
		\animategraphics[autoplay,loop,height = .173\linewidth, width = .173\linewidth]{5}{images/teaser/result/}{1}{39}&
		\includegraphics[height = .173\linewidth, width = .260\linewidth]{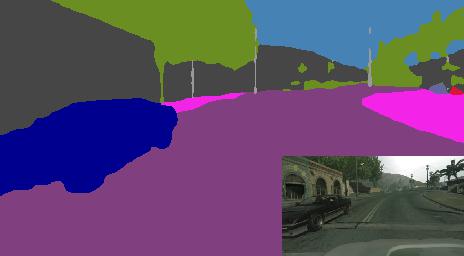} &\\
		{\footnotesize (a) artistic }& {\footnotesize (b) photo-realistic } & { \footnotesize (c) video }& { \footnotesize (d) domain adaption }
	\end{tabular}
	\vspace{-2mm}
		\caption{\footnotesize Applications of the proposed algorithm. (a) Artistic style transfer. (b) Photo-realistic style transfer. (c) Video style transfer (Click on the image to see animations using Adobe Reader). (d) Domain adaptation. Style thumbnails are shown in bottom right corners for the first three tasks, game image and transferred image are shown in bottom right corner for the last task.%
	}
	\label{fig:teaser}
\end{figure*}
\vspace{-2mm}

\begin{abstract}
Given a random pair of images, an arbitrary style transfer method extracts the feel from the reference image to synthesize an output based on the look of the other content image.
Recent arbitrary style transfer methods transfer second order statistics from reference image onto content image via a multiplication between content image features and a transformation matrix, which is computed from features with a pre-determined algorithm.
These algorithms either require computationally expensive operations, or fail to model the feature covariance and produce artifacts in synthesized images.
Generalized from these methods, in this work, we derive the form of transformation matrix theoretically and present an arbitrary style transfer approach that learns the transformation matrix with a feed-forward network.
Our algorithm is highly efficient yet allows a flexible combination of multi-level styles while preserving content affinity during style transfer process.
We demonstrate the effectiveness of our approach on four tasks: artistic style transfer, video and photo-realistic style transfer as well as domain adaptation, including comparisons with the state-of-the-art methods. 

\end{abstract}

\vspace{-5mm}
\section{Introduction}
\vspace{-2mm}
A style transfer method takes a content and a style image as inputs to synthesize an image with the look from the former and feel from the latter.
In recent years, numerous style transfer methods have been developed.
The method by Gatys et al.~\cite{gatys2015neural} iteratively optimizes content and style reconstruction losses between the target image and input images. 
To reduce the computational cost, 
a few approaches have since been developed based on feed-forward networks~\cite{johnson2016perceptual,ulyanov2016texture}.
However, these approaches cannot generalize to arbitrary style images with a single network.

For universal style transfer, a number of methods explore the second order statistical transformation from reference image onto content image via a linear multiplication between content image features and a transformation matrix~\cite{huang2017adain,WCT-NIPS-2017,li2018photoWCT}.
%
The AdaIn method is developed~\cite{huang2017adain} 
by matching the mean and variance of intermediate features between content and style images.
The WCT~\cite{WCT-NIPS-2017} algorithm further explores the covariance instead of the variance, by
embedding a whitening and coloring process within a pre-trained encoder-decoder module.
However, these approaches directly compute these matrices from features.
On one hand, they do not explore the general solutions to this problem. 
On the other hand, such matrix computation based methods can be expensive when the feature has a large dimension.

Generalized from these two methods~\cite{huang2017adain,WCT-NIPS-2017}, in this work we present theoretical analysis of such linear style transfer frameworks. 
We derive the form of transformation matrix and draw connections between this matrix to the style reconstruction objective (squared Frobenius norm of the difference between Gram matrices) widely used in style transfer~\cite{gatys2015neural,johnson2016perceptual,ulyanov2016texture,huang2017adain}. 
Based on the analysis, we learn the transformation matrix with two light-weighted CNNs.
We show that the learning-based transformation matrix can be controlled by different levels of style losses, and is highly efficient.
The contributions of this work are:
\vspace{-2mm}
\begin{itemize}
    \item We present general solutions to the linear transformation approach and show that optimizing it can also minimize the style reconstruction losses being used by the majority existing approaches.
    \item We propose a transformation matrix learning approach that is highly efficient ($\sim140$ fps), flexible (allows combination of multi-level styles in a single transformation matrix) and preserves content affinity during transformation process.
    \item With the flexibility, we show that the proposed method can be applied to many tasks, including but not limited to: artistic style transfer, video and photo-realistic style transfer, as well as domain adaptation, with comparisons to state-of-the-art methods.
\end{itemize}

\section{Related Work}
Style transfer has been studied in computer vision with different emphasis for more than two decades.
Early methods tackle this problem by non-parametric sampling~\cite{efros2001image}, non-photorealistic rendering~\cite{kyprianidis2013state} and image analogy~\cite{hertzmann2001image}. 
Most of these approaches rely on finding low-level image correspondence and do not capture high-level semantic information well.
Recently, Gatys et al.~\cite{gatys2015neural} propose an algorithm that matches statistical information between features of content and style images extracted from a pre-trained convolutional neural network. 
Numerous methods have since been developed to improve artistic style transfer quality~\cite{li2016combining,Wang2017MultimodalTA,wilmot2017stable}, add user control~\cite{wilmot2017stable,gatys2017controlling}, introduce more diversity~\cite{ulyanov2017improved,li2017diversified} or include semantic information~\cite{champandard2016semantic,frigo2016split}. 

One main drawback with the method by Gatys et al.~\cite{gatys2015neural} is the heavy computational cost due to the iterative optimization process. 
Fast feed-forward approaches~\cite{johnson2016perceptual,ulyanov2016texture,li2016precomputed} address this issue by training a feed-forward neural network that minimizes the same feature reconstruction loss and style reconstruction loss as~\cite{gatys2015neural}. 
However, each feed-forward network is trained to transfer exactly one fixed style.  
Dumoulin et al.~\cite{dumoulin2016learned} introduce an instance normalization layer that allows 32 styles to be represented by one model, and Li et al.~\cite{li2017diversified} encode 1,000 styles by using a binary selection unit for image synthesis.
Nevertheless these models are not able to transfer an arbitrary style onto a content image.

Recently, several methods \cite{BMVC-2017, huang2017adain,sheng2018avatar} are proposed to match mean and variance of content and style feature vectors to transfer an arbitrary style onto a content image. 
However, their methods do not model the covariance of the features and could not synthesize satisfying results in certain circumstances. 
Li et al.~\cite{WCT-NIPS-2017} resolve this problem by applying a whitening and coloring scheme with pre-trained image reconstruction auto-encoders.
However, this method is costly due to the need of large matrix decomposition, and auto-encoder cascading in order to combine multiple levels of styles.
Shen et al.\cite{shen2018style} proposes to train a meta network that generates a 14 layer network for each content/style image pair -- with expensive cost of extra memory for each content/style image pair, it does not explicitly model the covariance shift.

Different from existing approaches, we replace the transformation matrix computation in the intermediate layers with feed-forward networks, which can be more flexible and efficient.
We demonstrate the proposed algorithm can be effectively applied in four different tasks: artistic style transfer, video and photo-realistic style transfer, as well as domain adaptation. 

\begin{figure*}[t]
\centering
\includegraphics[width=12cm,scale=0.5]{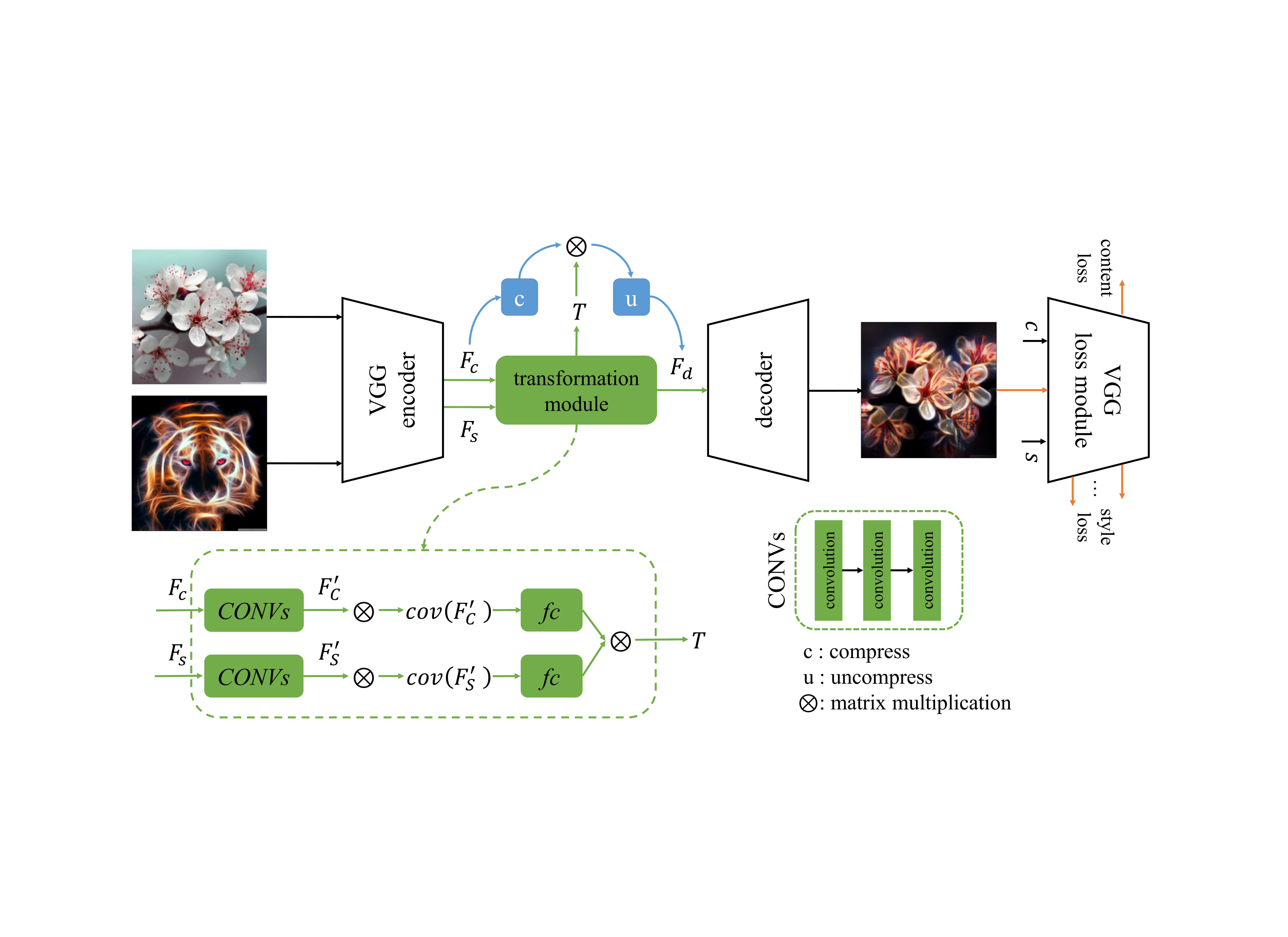}
\caption{Overview of the proposed method. Our model contains a pre-trained encoder and decoder, a loss module, a transformation module with the compress/uncompress blocks.
%
Only the transformation module, as well as the pair of compress and uncompress blocks are learnable, while all the others are fixed (black).
We use \textit{orange} arrows to denote the losses and the ``T'' to denote the point in which style is transformed (see Section~\ref{Architecture} for technical details).
}\vspace{-3mm}
\label{fig:overview}
\end{figure*}

\section{Style Transfer by Linear Transformation}
\label{sec:algorithm}
The proposed model contains two feed-forward networks, a symmetric encoder-decoder image reconstruction module and a transformation learning module, as shown in  Figure~\ref{fig:overview}.
The encoder-decoder is trained to reconstruct any input image faithfully. 
It is then fixed and serves as a base network in the remaining training procedures.
The transformation learning module contains two small CNNs which takes features of the content and style images from the top of the encoder respectively, and outputs a transformation matrix $T$.
The image style is transferred through linear multiplication between the content features and the transformation matrix $T$ in the same layer.
We use a pre-trained and fixed VGG-19 network to compute style losses at multiple levels
and one content loss in a way similar to the prior work~\cite{johnson2016perceptual,huang2017adain}.
The proposed model is a pure feed-forward convolutional neural network, which is able to  transfer arbitrary styles efficiently ($\sim140$ fps)

\subsection{Objectives for Arbitrary Style Transfer}
\label{sec:obj}
We denote the feature map of the top-most encoder layer 
as $F(I)\in \mathcal{R}^{N\times C}$, where $I$ is an input image, 
$N$ is the number of pixels and $C$ is the number of channels.
For presentation clarity, the feature maps of a content and style image pair are denoted as row vectors 
$F_c$ and $F_s$.
We model the image style transfer problem as a linear transformation between the content feature $F_c$ and the learned matrix $T$, with the transformed feature vector being $F_d$.
We use $\phi_{s}$ to denote a ``virtual'' feature map that provides a desired level of style.
For several matrix computation based methods~\cite{WCT-NIPS-2017,huang2017adain}, $\phi_{s}=F_s$. 
In our work, $\phi_{s}$ has multiple choices, 
which is controlled via different configurations of style losses in the loss module during training. 
Thus, $\phi_{s}$ can be described as a nonlinear mapping of $F_s$ as $\phi_{s} = \phi(F_s)$.
We denote $\bar{F}$ as the vectorized feature map $F$ with zero mean.

The objective of learning the style transformation is equivalent to minimizing the difference of centered covariance between ${F_d}$ and ${\phi_{s}}$: 
%
\begin{eqnarray} 
\label{eq:goal}
\mathcal{L}_{style} = & \frac{1}{NC}\norm{{\bar{F_d}}{\bar{F_d}}^{\top}-\bar{\phi_s}\bar{\phi_s}^{\top}}_F^2 \\ \nonumber
s.t.&  \bar{F_d} = T\bar{F_c}
\end{eqnarray}

By substituting the linear constraint into $\mathcal{L}_{style}$, the minima is obtained when\vspace{-2mm}
\begin{equation}
T\bar{F_c}\bar{F_c}^{\top}T^{\top} = \bar{\phi_s}\bar{\phi_s}^{\top}
 \label{eq:T}
\end{equation}
%
The centered covariance of $\bar{F_c}$ is $cov\left( F_c\right) = \bar{F_c}\bar{F_c}^{\top} = V_cD_cV_c^{\top}$, 
and the corresponding singular value decomposition (SVD) is  $cov\left( \phi_{s}\right) = \bar{\phi_s}\bar{\phi_s}^{\top} = V_sD_sV_s^{\top}$.
%
%
It is easy to show that 
\vspace{-2mm}
	\begin{equation}
	T = \left( V_sD_s^{\frac{1}{2}}V_s^{\top}\right)U\left( V_cD_c^{-\frac{1}{2}}V_c^{\top}\right) 
	\label{eq:T2}
	\end{equation}
is one set of solutions to \eqref{eq:T} where $U\in \mathcal{R^{C\times C}}$ is a $C$-dimensional orthogonal group. 
In other words, $T$ is solely determined by the covariance of the content and style image feature vectors.
Given $T$, the transformed feature is obtained by $\bar{F_d}+\mathit{mean}(F_s)$, which simultaneously aligns to the mean and the covariance statistics of the target style, and thus models the style reconstruction being used by the majority existing approaches.
In the following, we discuss how to select a proper model for learning $T$.

\subsection{Model for learning transformation $T$.}
Given that $T$ is only conditioned on the content and style images, one feasible approach is to use a network that takes both images to directly output a $C\times C$ matrix.
According to \eqref{eq:T2} where the terms of content and style are independent, we use two CNNs for the content/style inputs to isolate their interaction and make the structure more suitable.

%
The formulation in \eqref{eq:T} suggests three choices of input to the CNNs:
(i) images ($c$ and $s$), (ii) feature maps ($F_c$ and $F_s$), and (iii) covariance matrix ($cov\left( F_c\right) $ and $cov\left( F_s\right) $).
In this work, we use the third option where each CNN takes the covariance of feature vectors and outputs a $C\times C$ intermediate matrix. 
These two matrices are then multiplied to formulate the $T$ (see Figure~\ref{fig:overview}).
We explain the design choices in the following paragraph.

First, $T$ is irrelevant to the dimension of the input, e.g., a arbitrary region rather than the whole image needs to be transferred (e.g., photo-realistic stylization in Section~\ref{sec:exp}), when the covariance matrix is used. 
However, this property does not hold when the model input is the image or feature map.
For example, it is easy to prove that $T=\bar{\phi_s}U\bar{F_c}^{-1} $, another set of solutions, requires that the content and style feature input to the transformation module has the same dimensions.
Second, we need to use a fully-connected layer at the top of the two CNNs, since $T$ only contains global information -- in this case, using covariance matrix as the inputs is much more flexible, since using images or features indicates that all the input should have a fixed dimension under any circumstance.

Moreover, the above example for $T$ also suggests that using individual images or feature maps as model input leads to larger solution spaces, which potentially results in unstable local minima.
The ablation study in Section~\ref{sec:exp} shows that using a covariance matrix as model input leads to better results for general style transfer (see Figure~\ref{fig:struc_compare}).

\subsection{Efficient Model}\vspace{-1mm}
Our method achieves significant advantages in terms of computational efficiency (see Table~\ref{tab:time_compare}).
We show why the proposed model is more efficient and flexible as follows:


\noindent \textbf{Replacing matrix decomposition with a feed-forward network.}
The key to the speedup is that our model removes all time-consuming matrix computation on GPUs. 
Instead of using the SVD decomposition, e.g., in WCT~\cite{WCT-NIPS-2017}, which is not as GPU-friendly as CNN, we use a feed-forward network to approximate the matrix.
We show in our experiments that $T$ can be easily trained with a factorized CNN block and one fully-connected layer (the transformation module in Figure~\ref{fig:overview}). 

On the other hand, the AdaIn~\cite{huang2017adain} method partially addressed this issue by replacing the covariance matrix with the variance vector, but at the expense of image quality.
We show that while AdaIn only produces favorable results with a deeper auto-encoder, ours can achieve similar effects with much shallower auto-encoder, since the covariance shift is accurately modeled -- this brings a further speedup of the proposed algorithm.

\noindent \textbf{Learning multi-level style transfers via a single $T$.}
As analyzed in~\cite{gatys2015neural}, Gram matrices of features from different layers capture different details for style transfer. 
Gram matrices using features from lower layers (e.g., \textit{relu1\_1} and \textit{relu2\_1}) capture color and textures while those based on features from higher layers (e.g.  \textit{relu3\_1} and \textit{relu4\_1}) capture common patterns. 
%
Since the learnable $\phi_{s}$ is not necessarily equal to $F_s$, a single $T$ can express rich styles by using a combination of multiple style reconstruction losses in the loss module, e.g., via \{\textit{relu1\_1}, \textit{relu1\_2}, \textit{relu1\_3}, \textit{relu1\_4} \} in a way similar to \cite{huang2017adain,johnson2016perceptual} as shown in Figure~\ref{fig:loss_combi}.
This demonstrates a clear advantage over the matrix computation-based methods~\cite{WCT-NIPS-2017,huang2017adain}, in which the transform can only be computed via a certain layer in the encoder.

To address this issue, the WCT method \cite{WCT-NIPS-2017} transfers richer styles by cascading several auto-encoders with different layers, but at the expense of additional computational loads. 
In contrast, the proposed method only needs to use different loss networks (see the bottom row in Figure~\ref{fig:loss_combi}) -- it achieves the same goal without any computational overhead during the inference stage.

\subsection{Un-distorted Style Transfer}
\vspace{-1mm}
We show that the general linear transformation based style transfer methods enjoy the affinity preservation property, as introduced below.
By utilizing the property and a shallower auto-encoder, our method can be generalized to un-distorted style transfer.

\paragraph{\bf Affinity preserving for linear transformation models.} Affinity describes the pairwise relation of pixels, features, or other image elements. 
Aside from applying a learned style transformation, in some applications such as photo-realistic style transfer,the resulting images or videos also should not contain visually unpleasant distortions or artifacts. 
As such, preserving the affinity of content images, which indicates that the dense pairwise relations among pixels in the content image is preserved well in the result, is equally important for style transfer.
Many methods, such as guided filtering~\cite{he2013guided,liu2017spn}, matting Laplacian~\cite{levin2008closed,luan2017deep,li2018photoWCT} and non-local means~\cite{buades2005non}, aim to preserve the affinity of a guidance image, such that the output can be well-aligned to its input image structures (e.g., contours).

In principle, style transfer and manipulating affinity belong to different forms of matrix multiplications: 
the former is pre-multiplication while the latter is post-multiplication to the feature vector (see \eqref{eq:goal}).
%
We show that general linear transformation method is capable of preserving the feature affinity of the content image.
Given \eqref{eq:goal}, and the normalized affinity for a vectorized feature $F\in \mathcal{R^{N\times C}}$ as:
\begin{equation}
	\mathit{aff}\left(F \right)  = \bar{F}^{\top}cov\left( F\right) ^{-1}\bar{F}
	\label{eq:aff}
\end{equation}
It is easy to prove by the following equations that the affinities of $F_c$ and $F_d$ are equal to each other:
\begin{eqnarray}
	&\bar{F_d}^{\top}cov\left( \phi_s\right) ^{-1}\bar{F_d} = \bar{F_d}^{\top}\left( T\bar{F_c}\bar{F_c}^{\top}T^{\top} \right)^{-1} \bar{F_d} \\ \nonumber
	&= \bar{F_c}^{\top}T^{\top}\left( T\bar{F_c}\bar{F_c}^{\top}T^{\top} \right)^{-1}T\bar{F_c} =  \bar{F_c}^{\top}cov\left(F_c \right)^{-1} \bar{F_c}
\end{eqnarray}
%
%

\begin{figure}[t]
	\centering
	\begin{tabular}{c@{\hspace{0.005\linewidth}}c@{\hspace{0.005\linewidth}}c@{\hspace{0.005\linewidth}}c@{\hspace{0.005\linewidth}}c@{\hspace{0.005\linewidth}}c}
    	\includegraphics[height = .15\linewidth, width = .15\linewidth]{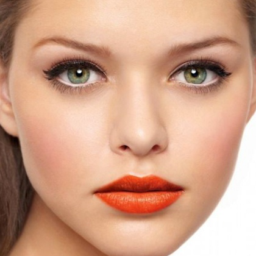} &
    	\includegraphics[height = .15\linewidth, width = .15\linewidth]{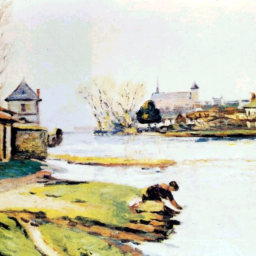} &
    	\includegraphics[height = .15\linewidth, width = .15\linewidth]{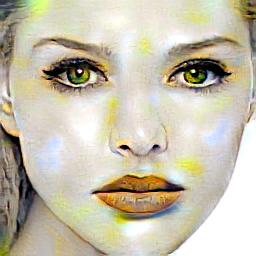} &
		\includegraphics[height = .15\linewidth, width = .15\linewidth]{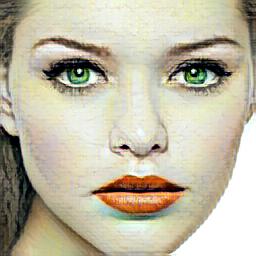} &
		\includegraphics[height = .15\linewidth, width = .15\linewidth]{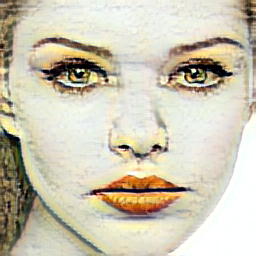} &\\
		{(a) \footnotesize Content }& {(b) \footnotesize Style }& { (c) \footnotesize WCT } & {  (d) \footnotesize AdaIn }& { (e) \footnotesize Ours }
	\end{tabular}
	\vspace{-2mm}
		\caption{Style transfer using a shallow auto-encoder.
	}
	\label{fig:3layer}
\end{figure}

\begin{figure*}[t]
		\centering
	\begin{tabular}{c@{\hspace{0.005\linewidth}}c@{\hspace{0.005\linewidth}}c@{\hspace{0.005\linewidth}}c@{\hspace{0.005\linewidth}}c@{\hspace{0.005\linewidth}}c@{\hspace{0.005\linewidth}}c@{\hspace{0.005\linewidth}}c@{\hspace{0.005\linewidth}}c}
		
		\includegraphics[height = .11\linewidth, width = .14\linewidth]{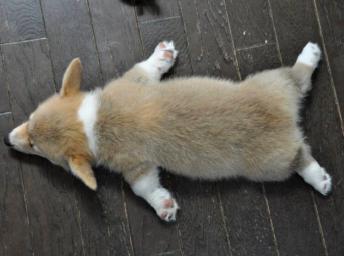} &
		\includegraphics[height = .11\linewidth, width = .14\linewidth]{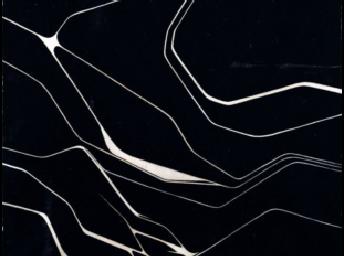} &
		\includegraphics[height = .11\linewidth, width = .14\linewidth]{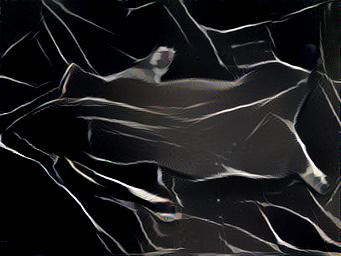} & 
		\includegraphics[height = .11\linewidth, width = .14\linewidth]{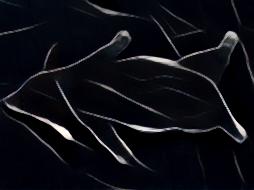} & 
		\includegraphics[height = .11\linewidth, width = .14\linewidth]{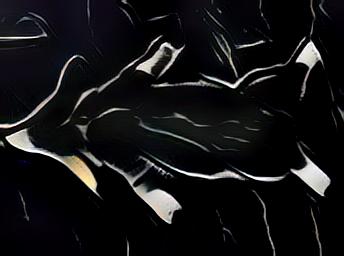} &
		\includegraphics[height = .11\linewidth, width = .14\linewidth]{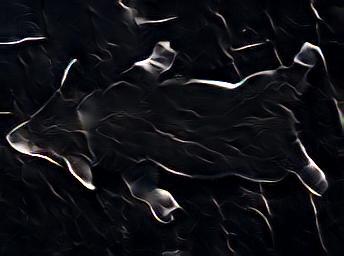} &
		\includegraphics[height = .11\linewidth, width = .14\linewidth]{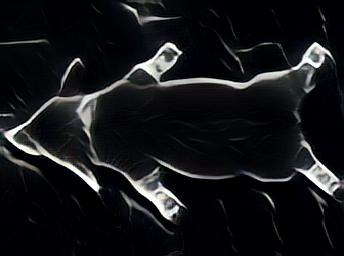} &  \\
		
		\includegraphics[height = .14\linewidth, width = .14\linewidth]{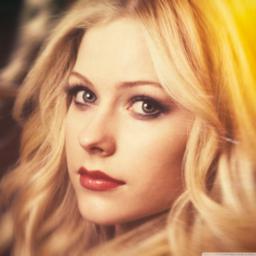} &
		\includegraphics[height = .14\linewidth, width = .14\linewidth]{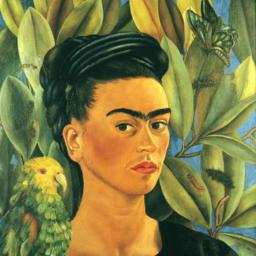} &
		\includegraphics[height = .14\linewidth, width = .14\linewidth]{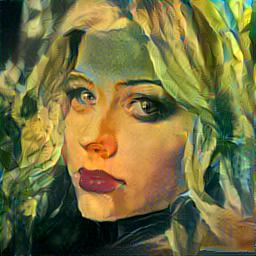} &
		\includegraphics[height = .14\linewidth, width = .14\linewidth]{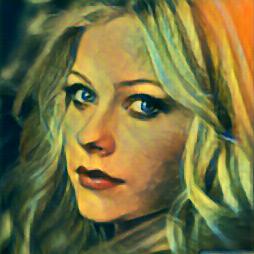} &
		\includegraphics[height = .14\linewidth, width = .14\linewidth]{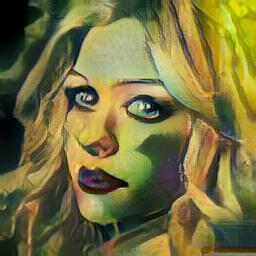} &
		\includegraphics[height = .14\linewidth, width = .14\linewidth]{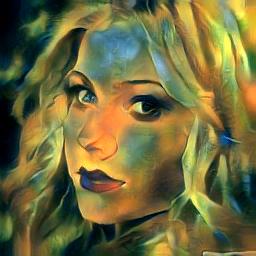} &
		\includegraphics[height = .14\linewidth, width = .14\linewidth]{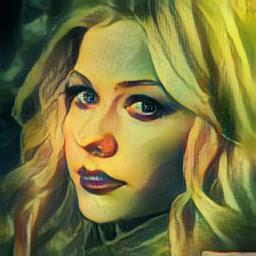} &   \\
		
		{ \footnotesize{Content} }& { \footnotesize{Style} } & {\footnotesize{Gatys~\cite{gatys2015neural}}} &{\footnotesize{Johnson~\cite{johnson2016perceptual}}}& { \footnotesize{Li~\cite{WCT-NIPS-2017}}} & {\footnotesize{ Huang~\cite{huang2017adain}} }&  { \footnotesize{Ours}} & \\
		
	\end{tabular}
	\vspace{-3mm}
	\caption{Affinity preserving comparison between our algorithm with~\cite{gatys2015neural,johnson2016perceptual,WCT-NIPS-2017,huang2017adain}. Our result has more similar affinity with content image than other methods. For instance, it faithfully preserves the contour of the dog (row 1) and the shadow of the face (row 2) during style transfer process.}
	\label{fig:affinity_compare}
\end{figure*}

\paragraph{\bf Utilizing of a shallower model.} 
%
%
There are three more factors that could cause distortion in the stylized image including 
(i) the decoder in which more nonlinear layers usually cause more distortions, (ii) $tr\left(\bar{\phi_s}\bar{\phi_s}^{\top}\right)$ that affects $F_d$, which is determined by how the loss is enforced (see similar analysis in~\cite{gupta2017characterizing}), and 
(iii) the spatial resolution of the top layer which determines at what scale the affinity is preserved.
Therefore, we propose to use a shallower model with up to \textit{relu3\_1} that better preserves the affinity.
Specifically, our shallower model can still express rich stylization when enforcing a relatively deeper style loss network (e.g., using up to \textit{relu4\_1} in the loss module), which however, can not be implemented in the same way both in AdaIn~\cite{huang2017adain} and WCT~\cite{li2018photoWCT}, as shown in Figure~\ref{fig:3layer}.
In other words, both AdaIn and WCT cannot effectively make use of the affinity preservation properties due to the limitation that they do not generalize well to a shallower network.
Aside from using a shallow encoder-decoder and a proper loss configuration, our method requires no extra manipulation, e.g., optical flow warpping~\cite{gupta2017characterizing} and matting Laplacian alignment~\cite{luan2017deep}, to render stylized images and videos.
We show that the stylized results contain fewer distortions and more stable sequences than existing methods~\cite{gatys2015neural,johnson2016perceptual} that are not based on the formulation in \eqref{eq:goal} (see Figure~\ref{fig:video_compare} and~\ref{fig:affinity_compare}).
Note that for a shallower encoder, the corresponding $F_c$ (e.g., \textit{relu2\_1}) is not expressive enough to transfer more abstract styles.

\vspace{-2mm}


\section{Network Implementation}
\vspace{-2mm}
\label{Architecture}
We describe the main modules in the proposed model as shown in Figure~\ref{fig:overview}.

\vspace{1mm}
\noindent \textbf{Encoder-decoder module.} 
Our encoder contains the first few layers from the VGG-19 model~\cite{Simonyan14c} pre-trained on the ImageNet dataset~\cite{imagenet_cvpr09}, and a symmetric decoder. 
%
We train the decoder on MS-COCO~\cite{lin2014microsoft} from scratch to reconstruct any input image.
This module is then fixed throughout the rest of the network training procedure.
We show various design options of the encoder-decoder architectures with respect to different applications in Section~\ref{sec:exp}
where the differences lie in model depth. 

\vspace{1mm}
\noindent \textbf{Transformation module.}
Our transformation module comprises of two CNNs, each of which takes either content or style feature as input, and outputs a transformation matrix respectively.
These two transformation matrices are then multiplied to
yield the final transformation matrix $T$. 
We produce the transferred feature vector $F_d$ by multiplying a content feature vector $F_c$ with $T$, then feed $F_d$ into the decoder to generate the stylized image. 

Within each CNN in the transformation module, as discussed in Section~\ref{sec:algorithm}, we compute transformation matrix from feature covariance to handle images of any size during inference. In order to learn a nonlinear mapping from feature covariance to transformation matrix, a group of fully-connected layers are preferred. 
However, this leads to a large memory requirement and model size since the covariance matrix usually has high dimensions, i.e., $512\times 512$ in \textit{relu4\_1}. 
Thus, we factorize the model by first encoding the input features to reduced dimensions (e.g., $512\rightarrow 32$) via three consecutive convolutional units (denoted as \enquote{CONVs} in Figure~\ref{fig:overview}), where each is equipped with a $3\times 3$ \textit{conv} and a \textit{relu} layer.
The covariance matrix of the encoded feature is then fed into one \textit{fc} unit to produce the transformation matrix.
We further adopt a pair of convolutional layers to ``compress'' the content feature, and ``uncompress'' the transformed feature to the corresponding dimensions.
Overall, our transformation module is compact, efficient and easy to converge for any combination of styles.

\vspace{1mm}
\noindent \textbf{Loss module.} 
Similar to existing methods~\cite{gatys2015neural,johnson2016perceptual,ulyanov2016texture}, we adopt a pre-trained VGG-19 as our loss network. 
We compute the style loss at different layers with the squared Frobenius norm of the difference between the Gram matrices, and the content loss as the Frobenius norm between content features and transferred features:
%
\begin{equation} 
\label{eq:Lcontent}
\mathcal{L}_{content} = \norm{{F_d}-{F_c}}^2 \\ 
\end{equation}
We train all models using features from \textit{relu4\_1} to compute the content loss.
The final loss is a weighted sum of content loss ${L}_{content}$ and style loss ${L}_{style}$: $\mathcal{L} = {L}_{content} + \lambda {L}_{style}$.
We present detailed analysis for the combination of style losses in Sec.~\ref{sec:exp}. 

\begin{figure*}[t]
\centering
\includegraphics[width=\textwidth]{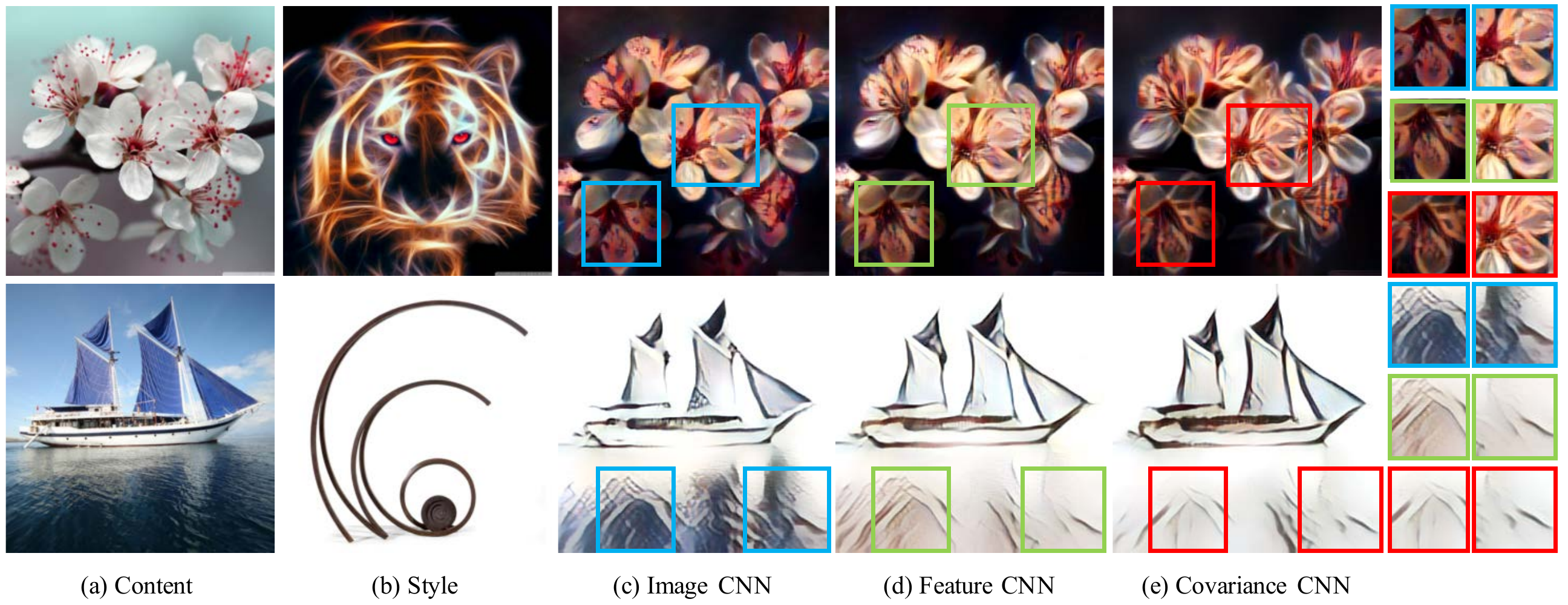}
\caption{Results from different model inputs. The CNN in (c) takes content/style image as input, the CNN in (d) takes content/style feature as input, and the CNN in (e) takes content/style feature as input but feeds the covariance matrix of encoded feature into the final fully connected layer (see Figure~\ref{fig:overview} and Section~\ref{sec:exp} for details).}
\label{fig:struc_compare}
\end{figure*}

\section{Experiment Results}
\label{sec:exp}
We discuss the experimental settings
and present ablation studies to understand how the main modules of the proposed algorithm contribute. 
We evaluate the proposed algorithm against the state-of-the-art methods on three style transfer tasks (artistic style transfer, video style transfer and  photo-realistic style transfer) and domain adaptation.

\subsection{Experimental Settings}
We use the MS-COCO dataset~\cite{lin2014microsoft} as our content images and the WikiArt database~\cite{wikiart} as our styles.  
Both datasets contain roughly 80,000 images. 
We keep the image ratio and rescale the smaller dimension of each training image to 300 pixels. 
We then apply patch-based training by randomly cropping a region of 256$\times$256 pixels from it as one training sample and randomly flip this sample with the probability of $0.5$.
On the inference stage, our model is able to handle any input size for both content and style images, as discussed in Section~\ref{sec:algorithm}.

We train our network using the Adam solver~\cite{kingma2014adam} with a learning rate of $10^{-4}$ and a batch size of 8 for $10^{5}$ iterations. 
The training roughly takes 2 hours on a single Titan XP GPU for each model in our Pytorch~\cite{paszke2017automatic} implementation.
The source code, trained models and real-time demos will be made available to the public.

\subsection{\bf Ablation Studies}
\noindent \textbf{Architecture.} 
We discuss in Section~\ref{sec:obj} that for learning $T$, using two separate CNNs should be more suitable than sharing a single CNN for content and style images, according to \eqref{eq:T2}.
To verify this, we train a single CNN to output the transformation matrix, which takes both content and style images as  inputs.
%
However, this model does not converge during training, which is consistent with our discussion. 
%

\begin{figure*}[t]
\centering
\includegraphics[width=\textwidth]{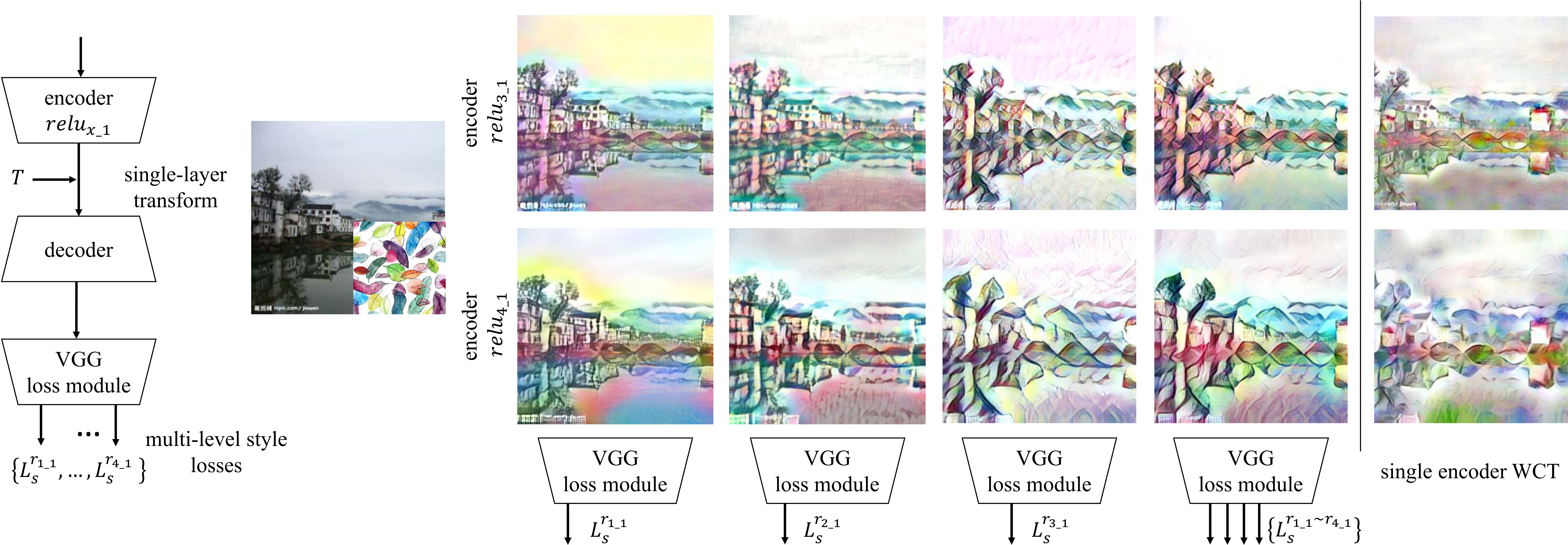}
\vspace{-3mm}
\caption{Our model can flexibly combine style losses at different levels. %
We use same style loss (shown in the last row) to train models in each column. Row 1 and row 2 show images synthesized by transferring content features from \textit{relu3\_1} and \textit{relu4\_1} respectively. The last column shows results by single layer WCT (relu3\_1 on the top and relu4\_1 at the bottom).}
\label{fig:loss_combi}
\vspace{-2mm}
\end{figure*}

\vspace{1mm}
\noindent \textbf{Inputs to the learning transformation module.} 
We discuss in Section~\ref{sec:algorithm} that the formulation in~\eqref{eq:T} suggests three input choices to CNNs in the transformation module.
We implement three models for the three different input choices, with the same architectures as the proposed model, except for the image CNN, which utilizes five convolutional units instead of three.
Note that both image and feature CNN do not have the matrix multiplication between the ``CONVs'' and the ``fc'' modules, as shown in Figure~\ref{fig:overview}.
We show comparisons in Figure~\ref{fig:struc_compare}.
In general, the image CNN does not produce faithful stylized results, e.g., the patches in the second row of the Figure~\ref{fig:struc_compare}(c) still present color from the original content image.
On the other hand, the covariance CNN (our proposed model) presents more similar stylized details (see patches in the close-ups in Figure~\ref{fig:struc_compare}(e)) than all the other methods.

\noindent \textbf{Combining multi-level style losses.} 
Features from different layers capture different scales of style. 
For instance, lower-layer features capture simpler styles such as color while higher-layer features are able to capture more complex styles such as textures or patterns.
We show our algorithm allows flexible combination of multiple levels of styles within a single transformation matrix $T$, by making use of different style reconstruction losses in the loss module. 
We apply two types of autoendocders -- encoders up to $relu3\_1$ and $relu4\_1$ in the VGG-19 model, and equip them with different style reconstruction losses -- single loss layer on $relu1\_1$, $relu2\_1$, $relu3\_1$, and multiple loss layers $\lbrace{relu1\_1, \cdots, relu4\_1\rbrace}$, as shown in Figure~\ref{fig:loss_combi}.

The results show that both transferring content features of lower layers (\textit{relu3\_1} in row 1) and using one single style reconstruction loss from lower layers (\textit{relu1\_1, relu2\_1} in column 2 and 3) lead to more photo-realistic visual effect. 
On the other hand, using style reconstruction loss from higher layers (e.g., column 3 and 4 of the Figure~\ref{fig:loss_combi}) produces more stylized images.
Specifically, the auto-encoder with the bottleneck as $relu3\_1$ (the first column in Figure~\ref{fig:loss_combi}) produces more un-distorced results for all types of style losses, than those with the bottleneck as $relu4\_1$ (the second column in Figure~\ref{fig:loss_combi}).

We also show stylized results by single-encoder WCT in the last column of  Figure~\ref{fig:loss_combi} (\textit{relu3\_1} on top and \textit{relu4\_1} on bottom),
where neither of them show satisfying style transfer result (e.g., the color and the texture is not well aligned, and the edges are blurry).
In contrast, our method produces visually pleasant style transfer results for richer styles, as shown in the fourth column in Figure~\ref{fig:loss_combi}, where no cascading of modules is required in contrast to the WCT method.
Thanks to the flexibility of our algorithm, we are able to choose different settings for different tasks without any computational overheads.
Specifically, we transfer content image features from \textit{relu4\_1} layer for artistic style transfer, while for photo-realistic and video style transfer, we transfer content image features from \textit{relu3\_1}  to avoid distortions and instabilities. For all tasks, we compute style reconstruction losses using features from \textit{relu1\_1}, \textit{relu2\_1}, \textit{relu3\_1} and \textit{relu4\_1} to capture different scales of style.

\begin{figure*}[t]
	\centering
	\begin{tabular}{c@{\hspace{0.005\linewidth}}c@{\hspace{0.005\linewidth}}c@{\hspace{0.005\linewidth}}c@{\hspace{0.005\linewidth}}c@{\hspace{0.005\linewidth}}c@{\hspace{0.005\linewidth}}c@{\hspace{0.005\linewidth}}c}
		
		\includegraphics[height = .14\linewidth, width = .14\linewidth]{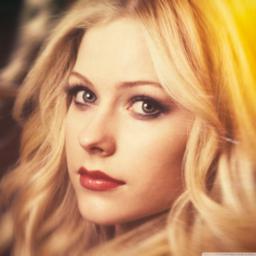} &
		\includegraphics[height = .14\linewidth, width = .14\linewidth]{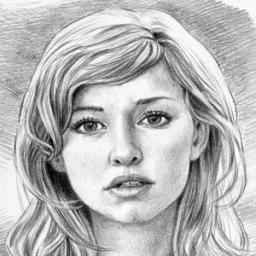} &
		\includegraphics[height = .14\linewidth, width = .14\linewidth]{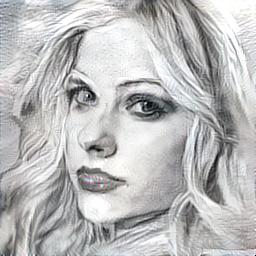} &
		\includegraphics[height = .14\linewidth, width = .14\linewidth]{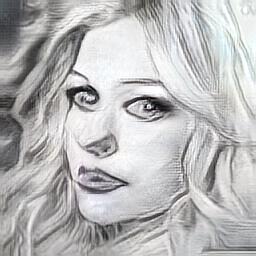} &
		\includegraphics[height = .14\linewidth, width = .14\linewidth]{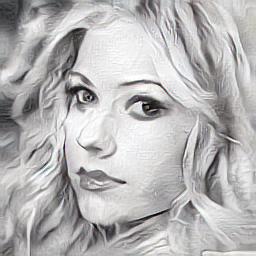} & 
		\includegraphics[height = .14\linewidth, width = .14\linewidth]{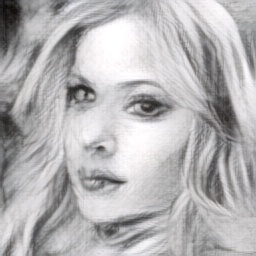} & 
		\includegraphics[height = .14\linewidth, width = .14\linewidth]{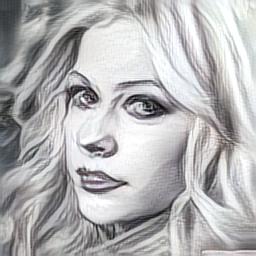} & \\
		
		\includegraphics[height = .14\linewidth, width = .14\linewidth]{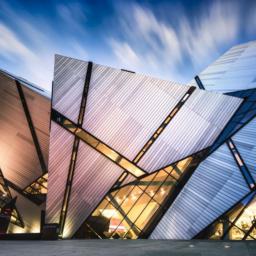} &
		\includegraphics[height = .14\linewidth, width = .14\linewidth]{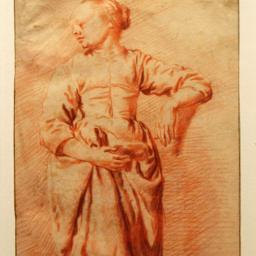} &
		\includegraphics[height = .14\linewidth, width = .14\linewidth]{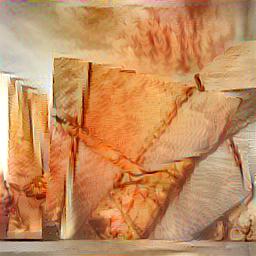} &
		\includegraphics[height = .14\linewidth, width = .14\linewidth]{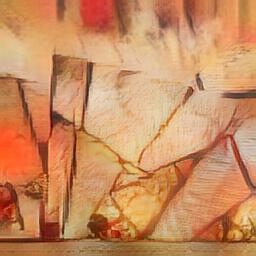} &
		\includegraphics[height = .14\linewidth, width = .14\linewidth]{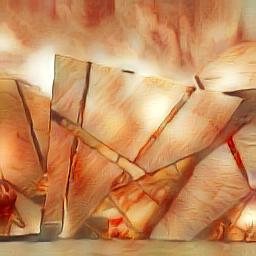} & 
		\includegraphics[height = .14\linewidth, width = .14\linewidth]{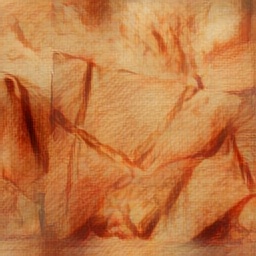} &
		\includegraphics[height = .14\linewidth, width = .14\linewidth]{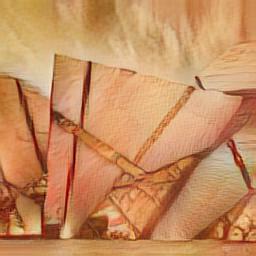} & \\
		
		\includegraphics[height = .14\linewidth, width = .14\linewidth]{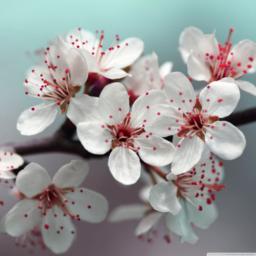} &
		\includegraphics[height = .14\linewidth, width = .14\linewidth]{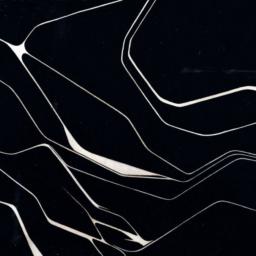} &
		\includegraphics[height = .14\linewidth, width = .14\linewidth]{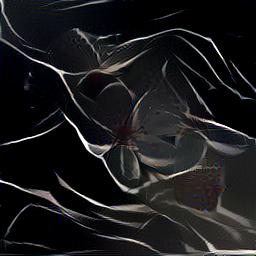} &
		\includegraphics[height = .14\linewidth, width = .14\linewidth]{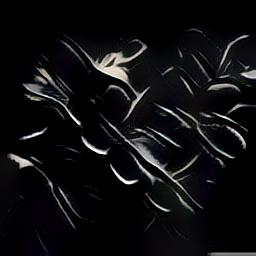} &
		\includegraphics[height = .14\linewidth, width = .14\linewidth]{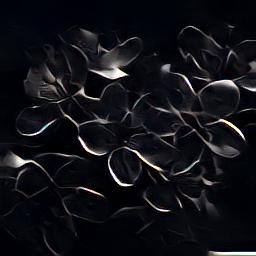} & 
		\includegraphics[height = .14\linewidth, width = .14\linewidth]{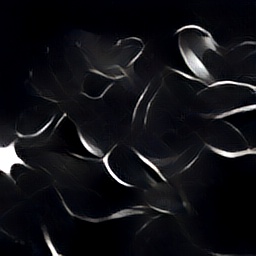} & 
		\includegraphics[height = .14\linewidth, width = .14\linewidth]{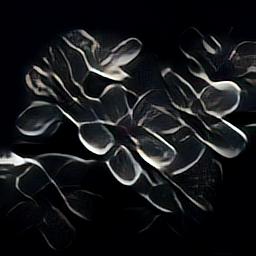} & \\
		
		\includegraphics[height = .10\linewidth, width = .14\linewidth]{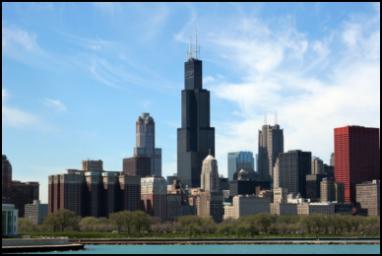} &
		\includegraphics[height = .10\linewidth, width = .14\linewidth]{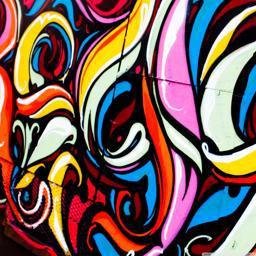} &
		\includegraphics[height = .10\linewidth, width = .14\linewidth]{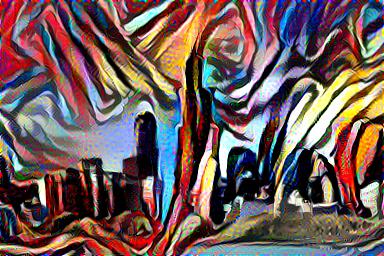} &
		\includegraphics[height = .10\linewidth, width = .14\linewidth]{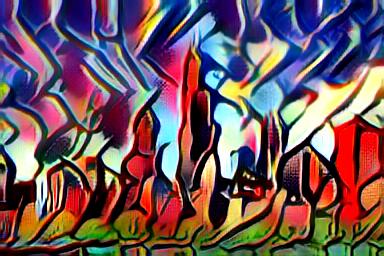} &
		\includegraphics[height = .10\linewidth, width = .14\linewidth]{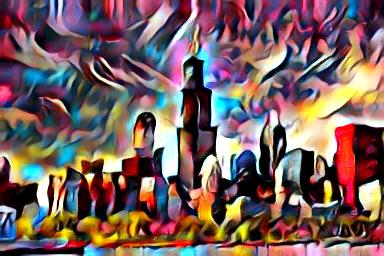} & 
		\includegraphics[height = .10\linewidth, width = .14\linewidth]{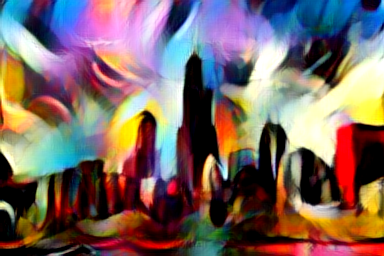} & 
		\includegraphics[height = .10\linewidth, width = .14\linewidth]{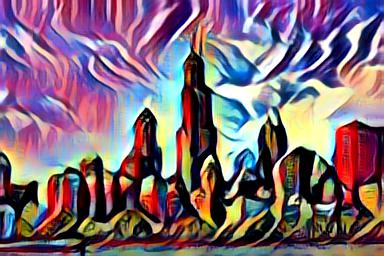} & \\
		
		\includegraphics[height = .14\linewidth, width = .14\linewidth]{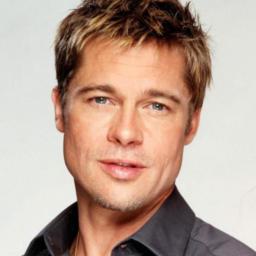} &
		\includegraphics[height = .14\linewidth, width = .14\linewidth]{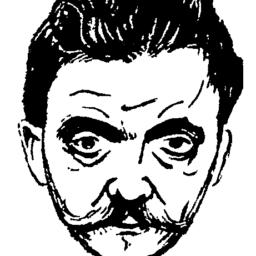} &
		\includegraphics[height = .14\linewidth, width = .14\linewidth]{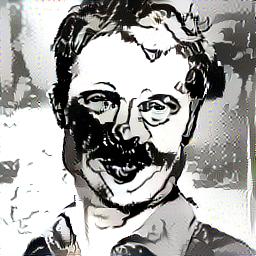} &
		\includegraphics[height = .14\linewidth, width = .14\linewidth]{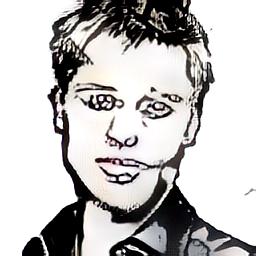} &
		\includegraphics[height = .14\linewidth, width = .14\linewidth]{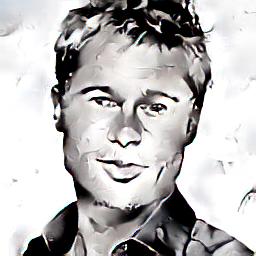} & 
		\includegraphics[height = .14\linewidth, width = .14\linewidth]{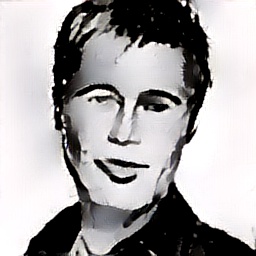} &
		\includegraphics[height = .14\linewidth, width = .14\linewidth]{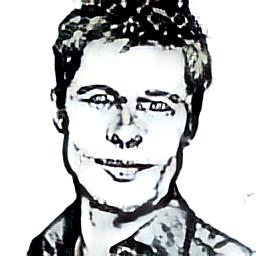} & \\
		\footnotesize{ Content } & \footnotesize{ Style } & \footnotesize{ Gatys~\cite{gatys2015neural}} & \footnotesize{Huang~\cite{huang2017adain}} & \footnotesize{Li~\cite{WCT-NIPS-2017}} & \footnotesize{Sheng~\cite{sheng2018avatar}} & \footnotesize{Ours }  & \\
	\end{tabular}
	\vspace{-3mm}
	\caption{Comparison between results by our style transfer algorithm and other methods. Our model is trained to transfer content features from \textit{relu4\_1} with style losses computed on \textit{relu1\_1}, \textit{relu2\_1}, \textit{relu3\_1}, \textit{relu4\_1} layer of VGG 19. All content images as well as style images have never been seen by our model during training process. More examples can be found in the supplementary materials.}
	\label{fig:general_compare}
\end{figure*}
\vspace{-2mm}

\begin{figure*}[h]
	\centering
	\begin{tabular}{c@{\hspace{0.005\linewidth}}c@{\hspace{0.005\linewidth}}c@{\hspace{0.005\linewidth}}c@{\hspace{0.005\linewidth}}c@{\hspace{0.005\linewidth}}c@{\hspace{0.005\linewidth}}c}
		
		{}&{ \textbf{Frames} }& { \textbf{Gatys~\cite{gatys2015neural}} } & { \textbf{Johnson~\cite{johnson2016perceptual}} }& { \textbf{Li~\cite{WCT-NIPS-2017}}}& { \textbf{Ours}}&\\
		
		{\rotatebox[origin=c]{90}{\textbf{Frame1}}}& \raisebox{-0.5\height}{\includegraphics[height = .182\linewidth, width = .182\linewidth]{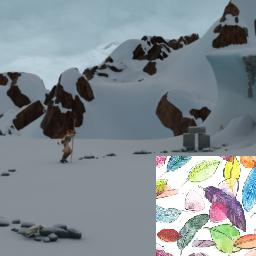}}& \raisebox{-0.5\height}{\includegraphics[height = .182\linewidth, width = .182\linewidth]{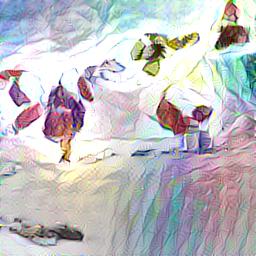}}& \raisebox{-0.5\height}{\includegraphics[height = .182\linewidth, width = .182\linewidth]{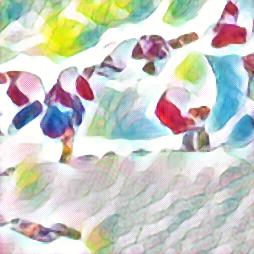}}& \raisebox{-0.5\height}{\includegraphics[height = .182\linewidth, width = .182\linewidth]{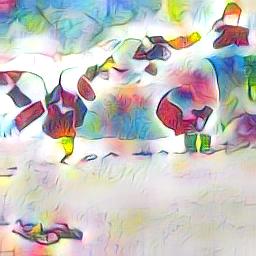}}& \raisebox{-0.5\height}{\includegraphics[height = .182\linewidth, width = .182\linewidth]{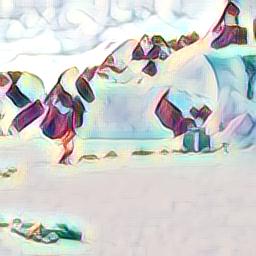}} \\[1.15cm]

		{\rotatebox[origin=c]{90}{\textbf{Frame5}}}& \raisebox{-0.5\height}{\includegraphics[height = .182\linewidth, width = .182\linewidth]{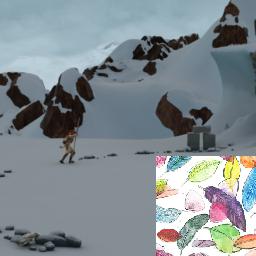}}& \raisebox{-0.5\height}{\includegraphics[height = .182\linewidth, width = .182\linewidth]{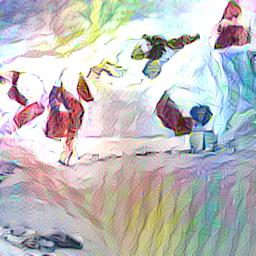}}& \raisebox{-0.5\height}{\includegraphics[height = .182\linewidth, width = .182\linewidth]{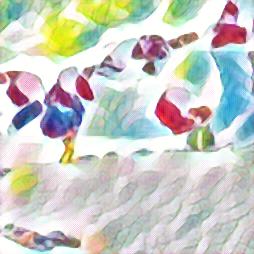}}& \raisebox{-0.5\height}{\includegraphics[height = .182\linewidth, width = .182\linewidth]{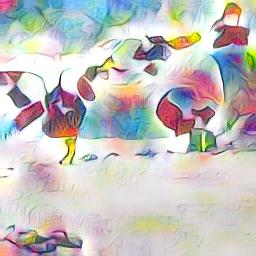}}& \raisebox{-0.5\height}{\includegraphics[height = .182\linewidth, width = .182\linewidth]{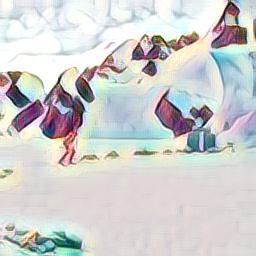}} \\[1.13cm]
		
		{\rotatebox[origin=c]{90}{\textbf{Difference}}}& \raisebox{-0.5\height}{\includegraphics[height = .157\linewidth, width = .182\linewidth]{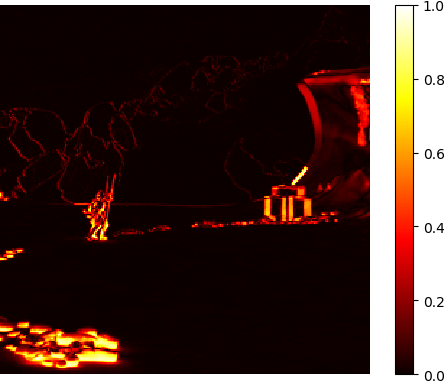}}& \raisebox{-0.5\height}{\includegraphics[height = .157\linewidth, width = .182\linewidth]{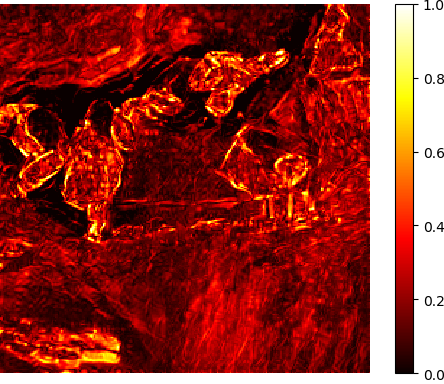}}& \raisebox{-0.5\height}{\includegraphics[height = .157\linewidth, width = .182\linewidth]{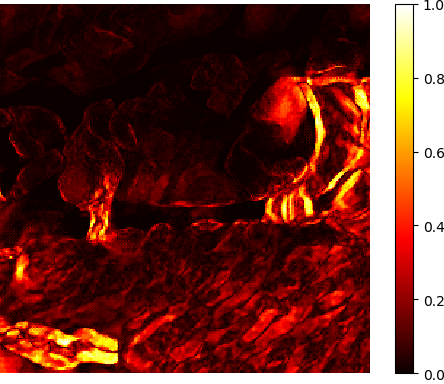}}& \raisebox{-0.5\height}{\includegraphics[height = .157\linewidth, width = .182\linewidth]{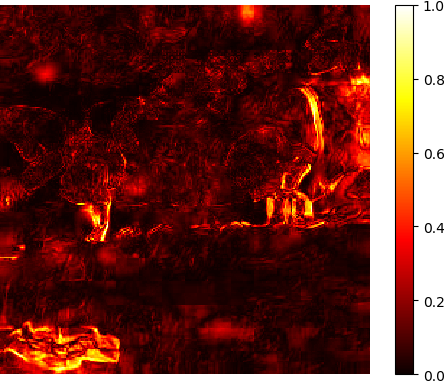}}& \raisebox{-0.5\height}{\includegraphics[height = .157\linewidth, width = .182\linewidth]{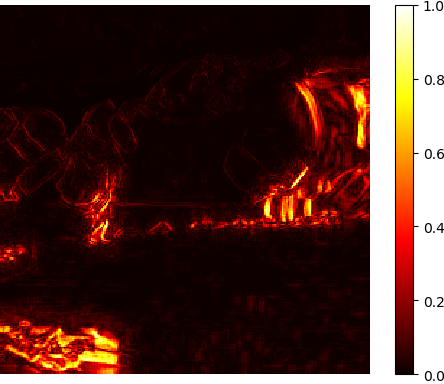}} \\
		
	\end{tabular}
	\vspace{-2mm}
	\caption{Video style transfer comparisons. The first row and second row show the first frame and fifth frame with corresponding transferred frames by evaluated method, and the last row shows the heat map of difference between these two frames. 
	%
	}
	\label{fig:video_compare}
\end{figure*}

\begin{figure}
    {
    \caption{\scriptsize User study of stylization effects on images and stability on videos.}\label{fig:user_study}}
    {\includegraphics[height = .71\linewidth, width = \linewidth]{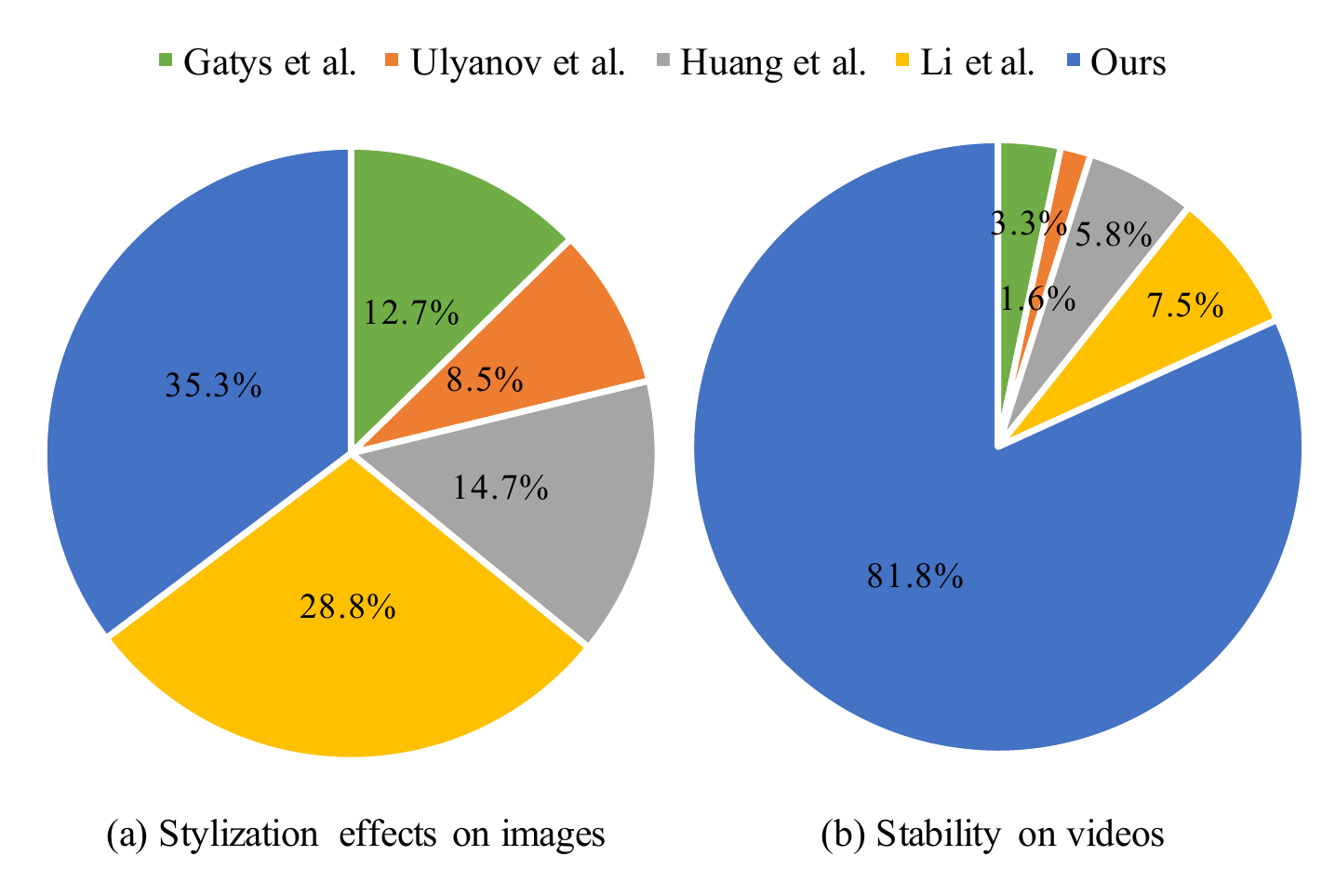}}
\end{figure}
\begin{table}
  {\caption{\scriptsize{Run time performance. ``N/A'' indicates the input cannot fit to 12GB GPU. Run time is measured in seconds using the original source code on a single Titan XP GPU. For WCT~\cite{WCT-NIPS-2017}, we use the version that cascades four different encoder-decoder modules because of its best performance and faster speed.}}\label{tab:time_compare}}
  {\scriptsize
  \begin{tabular}{llll}
  \hline
    {Image Size} & 256 & 512 & 1024\\ \hline
    {Ulyanov et al.~\cite{ulyanov2016texture}} & 0.013 & 0.028& 0.092\\
    \hline
      {Gatys et al.~\cite{gatys2017controlling}} & 16.51 &59.45&N/A\\
	   {Huang et al.~\cite{huang2017adain}} & 0.019 & 0.071 & N/A\\
	   {Li et al.~\cite{WCT-NIPS-2017}} &0.922 & 1.080 & N/A\\
	   {Ours (\textit{relu3\_1})} &0.007 & 0.025& 0.100\\
	   {Ours (\textit{relu4\_1})} &0.010 & 0.036 & 0.146\\
			 \hline
  \end{tabular}}
\end{table}

\subsection{Artistic Style Transfer}
\label{sec:general_compare}

We evaluate the proposed algorithm with three state-of-the-art methods for artistic style transfer: optimization based method~\cite{gatys2015neural}, fast feed-forward network~\cite{johnson2016perceptual} and feature transformation based approaches~\cite{huang2017adain,WCT-NIPS-2017}.

\noindent \textbf{Qualitative results.} We present stylized results of the evaluated methods in Figure~\ref{fig:general_compare} and more results in supplementary materials. 
The proposed algorithm performs favorably against the state-of-the-art methods. 
Although the optimization based method~\cite{gatys2015neural} allows arbitrary style transfer, it is computationally expensive due to iterative optimization process (see Table~\ref{tab:time_compare}). 
In addition, it encounters local minima issues which lead to over exposure images (e.g., row 5). 
The fast feed forward methods~\cite{johnson2016perceptual} improve the efficiency of optimization based method but it requires training one network for each style and requires adjusting style weights for best performance. \\
The AdaIn method presents an efficient solution to resolve arbitrary style transfer by matching mean and variance between style image and target image. However, it generates less appealing results (e.g., row 2, 3, 5) because it replaces covariance matrix with the variance vector, which is less effective in capturing second order statistics. 
The WCT method largely improves the visual quality of the transferred image (see column 4, 5 in Fig~\ref{fig:general_compare}) by modeling the divergence of the covariance matrices, but at the expense of inference speed (see Table~\ref{tab:time_compare}). 
In contrast, our method models the covariance-based transformation and shows favorable transferring results for arbitrary style (see the last column in Fig~\ref{fig:general_compare}), while still maintains the high efficiency (see Table~\ref{tab:time_compare}), and can apply to much wider applications due to its flexibility (see the following part in this section).

\vspace{1mm}
\noindent \textbf{User study.}
We conduct a user study to evaluate the proposed algorithm against the state-of-the-art style transfer methods~\cite{gatys2015neural,ulyanov2017improved,huang2017adain,WCT-NIPS-2017}. 
We use 6 content images and 40 style images to synthesize 240 images in total, and show 15 randomly chosen content and style combinations to each subject. 
For each combination, we present 5 synthesized images by each aforementioned method in a random order and ask the subject to select the most visually pleasant one.
We collect 540 votes from 36 users and show the percentage of votes for each method in Figure~\ref{fig:user_study}(a).
Overall, the proposed algorithm is favored among all evaluated methods. 

\vspace{1mm}
\noindent \textbf{Efficiency.}
Our method is efficient thanks to the feed-forward architecture. 
Table~\ref{tab:time_compare} shows the
run time performance of the proposed algorithm and other state-of-the-art methods on three input image scales: $256\times 256$, $512\times 512$, $1024\times 1024$. 
All methods listed in this table allow arbitrary style transfer except for the algorithm proposed by Ulyanov et al.~\cite{ulyanov2016texture} (row 1).
Even the slower variant of the proposed algorithm (trained to transfer content features from \textit{relu4\_1}) runs at 100 FPS and 27 FPS for $256\times 256$ and $512\times 512$ images, thereby making real-time style transfer feasible. 
Furthermore, our model has clear efficiency advantage over the optimization-based method~\cite{gatys2015neural} (about 3 orders of magnitude faster) as well as WCT~\cite{WCT-NIPS-2017} (about 2 orders of magnitude faster) and is comparable to the fast feed-forward methods~\cite{ulyanov2016texture,huang2017adain}.

\subsection{Video and Photo-realistic Style Transfer}
\label{sec:affinity}

Another crucial property of our algorithm is its ability to preserve content affinity during style transfer, when proper architectures are selected.
This is particularly important for tasks such as video and photo-realistic style transfer since these tasks prefer that both the global structures and the detailed contours in the content video or images can be preserved during style transfer process.
We discuss the results in the following part.

\vspace{1mm}
\noindent \textbf{Video style transfer.} 
Video style transfer is conducted between a content video and a style image in a frame-wise manner, by using a shallower auto-encoder with $relu3\_1$ as the bottleneck.
This computation can be more efficient than image based style transfer, since the transformation output in the style branch can be computed only once during initialization (see Figure~\ref{fig:overview}), and then directly applied to the rest of the frames. 
Since our approach can preserve the affinity of content videos, which are naturally consistent and stable, our stylized videos are also visually stable without the aid of any auxiliary techniques, such as optical flow warping~\cite{gupta2017characterizing}.
Figure~\ref{fig:video_compare} shows our direct frame-based style transfer results compared with existing style transfer methods~\cite{gatys2015neural,johnson2016perceptual,WCT-NIPS-2017}. 
To visualize stability of synthesized video clip, we draw heat maps of differences between two frames (i.e., row 3). 
The heat map difference by the proposed algorithm is closest to that of the original frames, which implies that our algorithm is able to preserve content affinity during style transfer process. 
To further evaluate the stability of our algorithm in video style transfer, we conduct a user study with 5 video clips synthesized frame-wise by our method and 4 state-of-the-art methods~\cite{gatys2015neural,ulyanov2016texture,huang2017adain,WCT-NIPS-2017}. 
For each group of videos, we ask each subject to select the most stable video clip. 
We collect 106 votes from 27 subjects and show percentage of votes for each method in Figure~\ref{fig:user_study}(b). 
Overall, the proposed  algorithm performs well against the other methods, which indicates the ability of our method to preserve affinity during style transfer.
More video transfer results can be found in the supplementary material.

\begin{figure*}[t]
	\centering
	\caption{\footnotesize{Photo-realistic style transfer results comparison. 
	Spatial mask is displayed at the right bottom corner of each content and style image. 
	\enquote{+Bilateral} in last column means results filtered by a bilateral filter after stylizaton.} 
	}
	\begin{tabular}{c@{\hspace{0.005\linewidth}}c@{\hspace{0.005\linewidth}}c@{\hspace{0.005\linewidth}}c@{\hspace{0.005\linewidth}}c@{\hspace{0.005\linewidth}}c}
		\includegraphics[height = .1\linewidth, width = .16\linewidth]{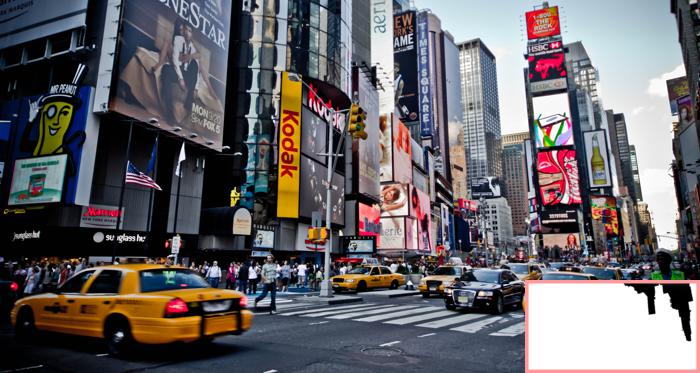} &
		\includegraphics[height = .1\linewidth, width = .16\linewidth]{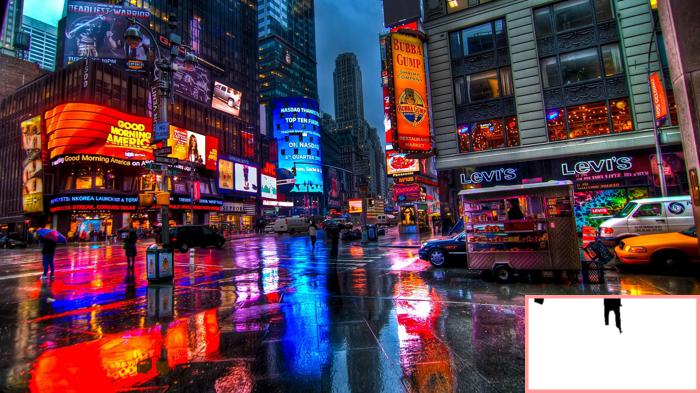} &
		\includegraphics[height = .1\linewidth, width = .16\linewidth]{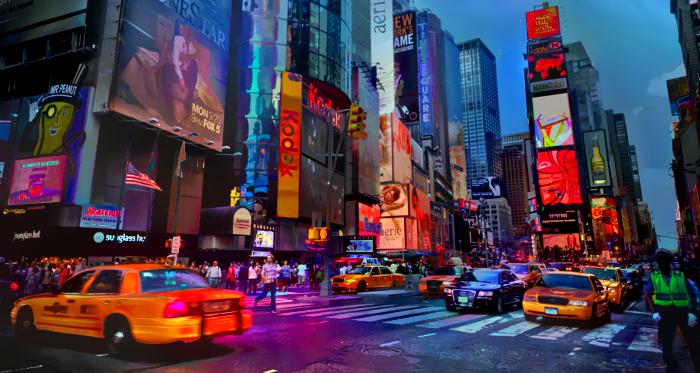} & 
		\includegraphics[height = .1\linewidth, width = .16\linewidth]{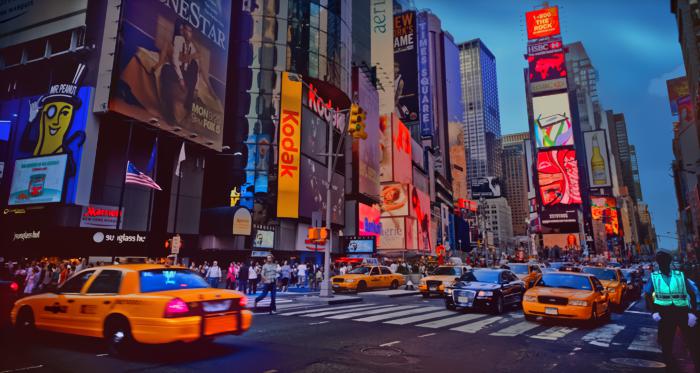} &
		\includegraphics[height = .1\linewidth, width = .16\linewidth]{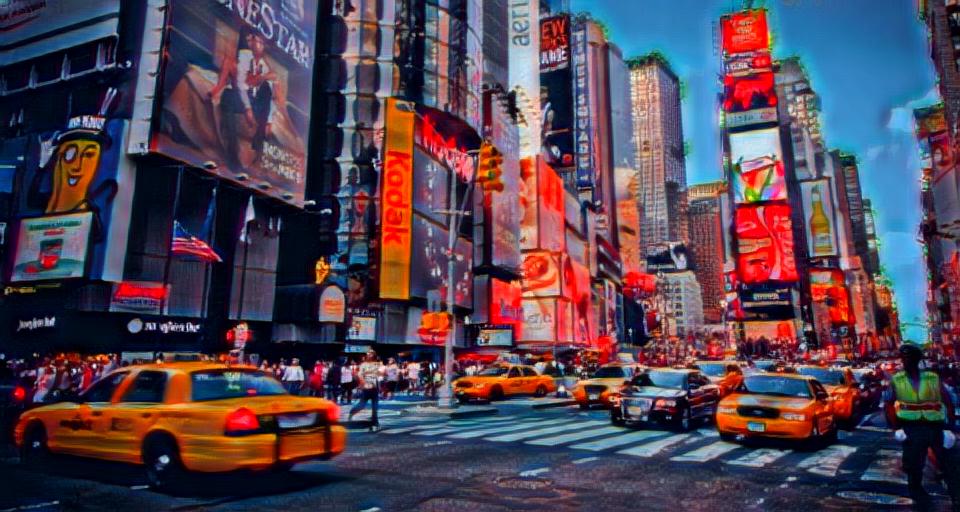} &
		\includegraphics[height = .1\linewidth, width = .16\linewidth]{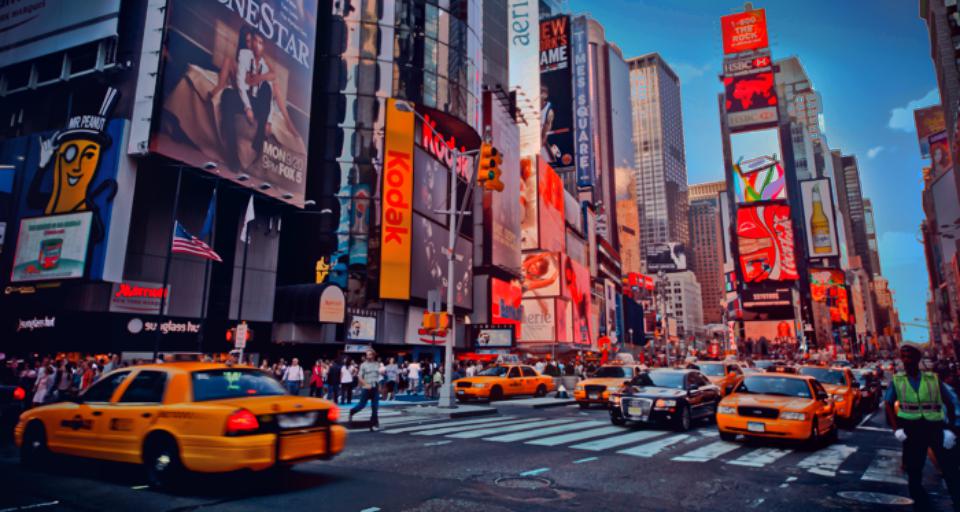}  \\
		
		\includegraphics[height = .1\linewidth, width = .16\linewidth]{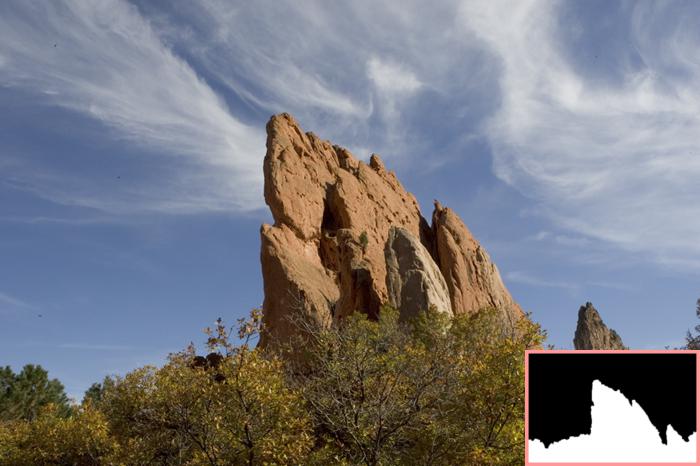} &
		\includegraphics[height = .1\linewidth, width = .16\linewidth]{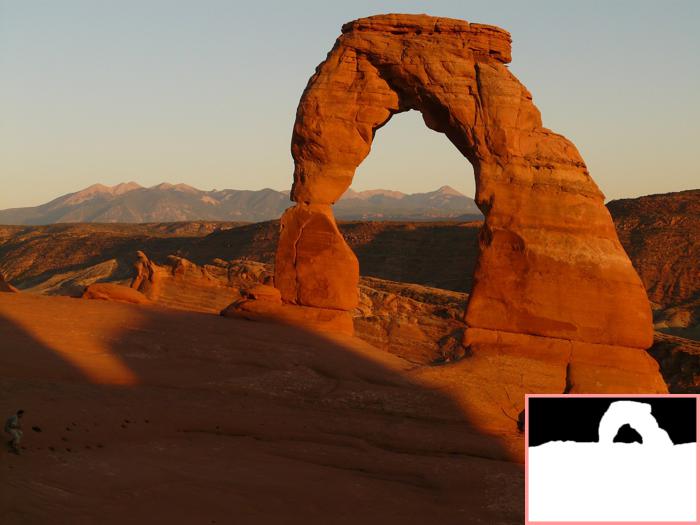} &
		\includegraphics[height = .1\linewidth, width = .16\linewidth]{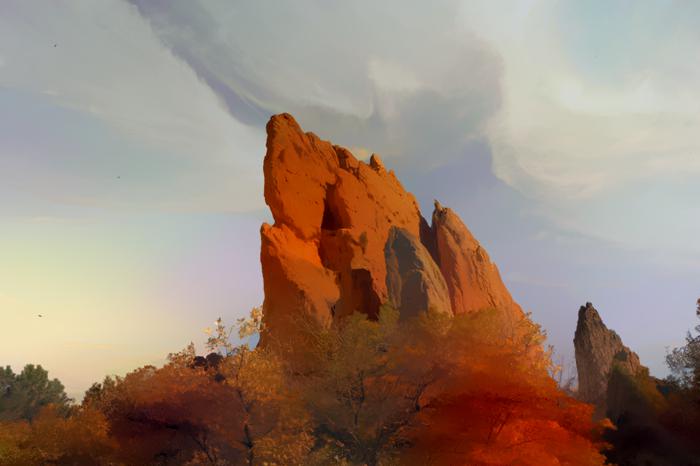} &
		\includegraphics[height = .1\linewidth, width = .16\linewidth]{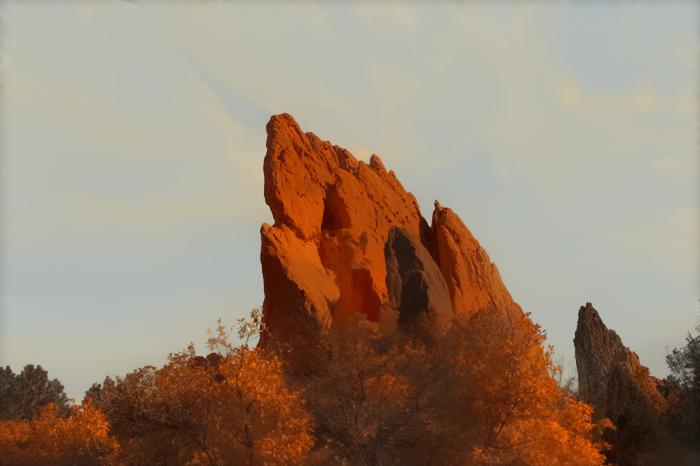} &
		\includegraphics[height = .1\linewidth, width = .16\linewidth]{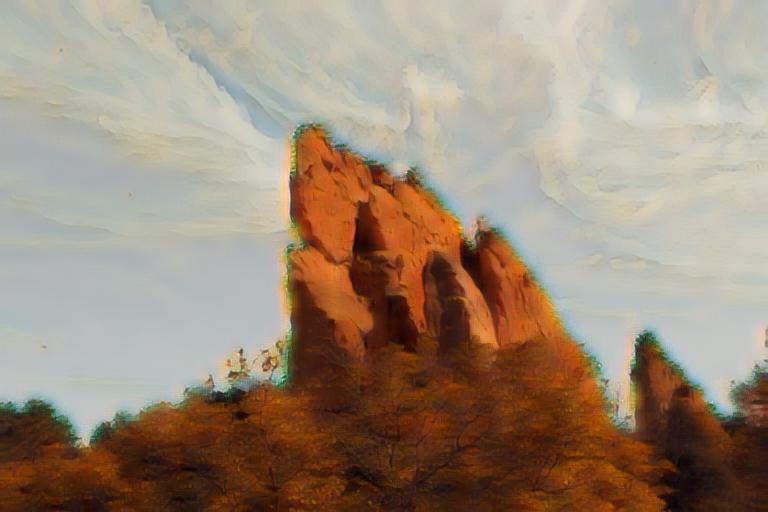} &
		\includegraphics[height = .1\linewidth, width = .16\linewidth]{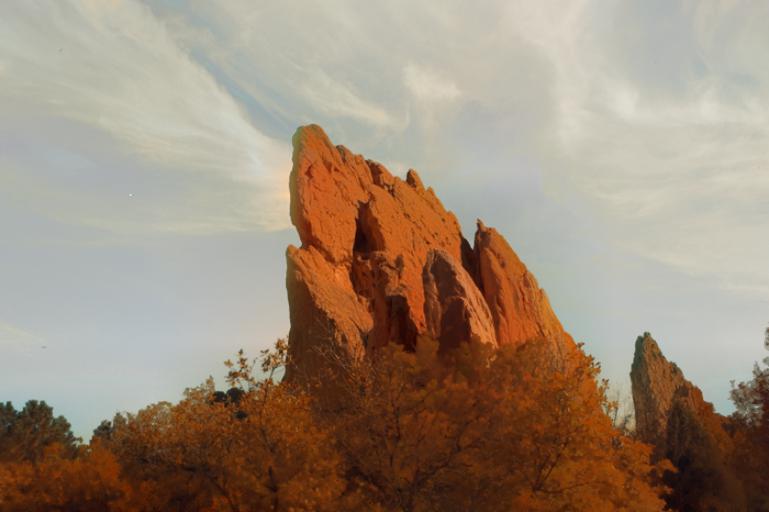}  \\
		
		\includegraphics[height = .13\linewidth, width = .16\linewidth]{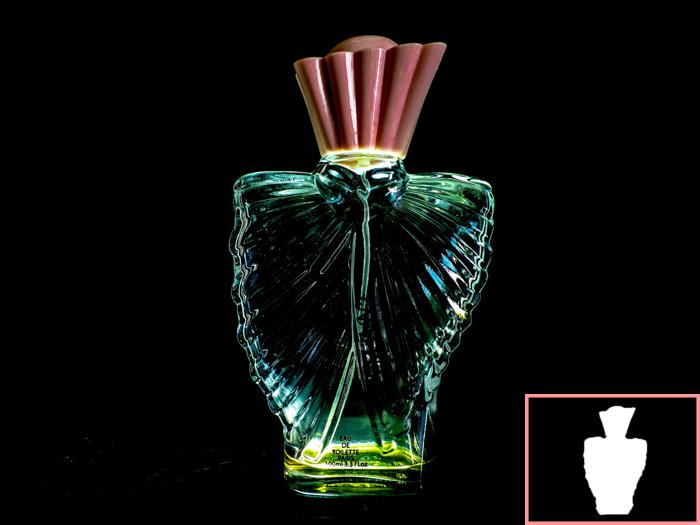} &
		\includegraphics[height = .13\linewidth, width = .16\linewidth]{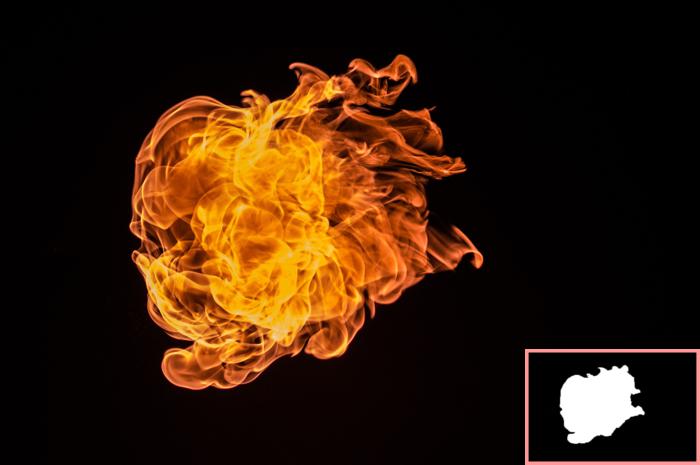} &
		\includegraphics[height = .13\linewidth, width = .16\linewidth]{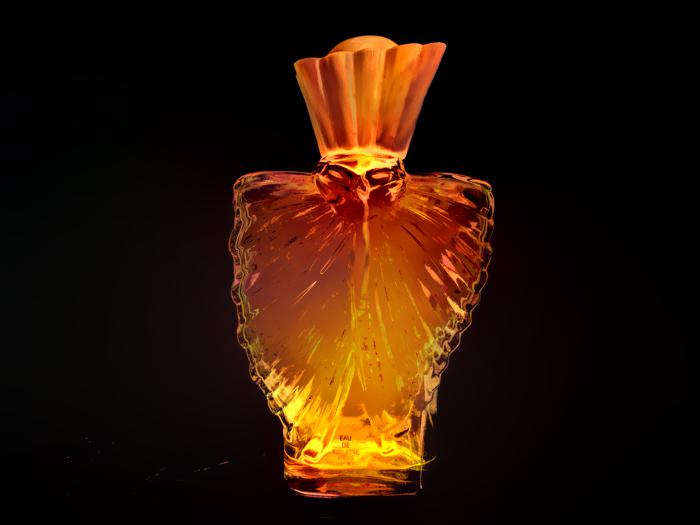} &
		\includegraphics[height = .13\linewidth, width = .16\linewidth]{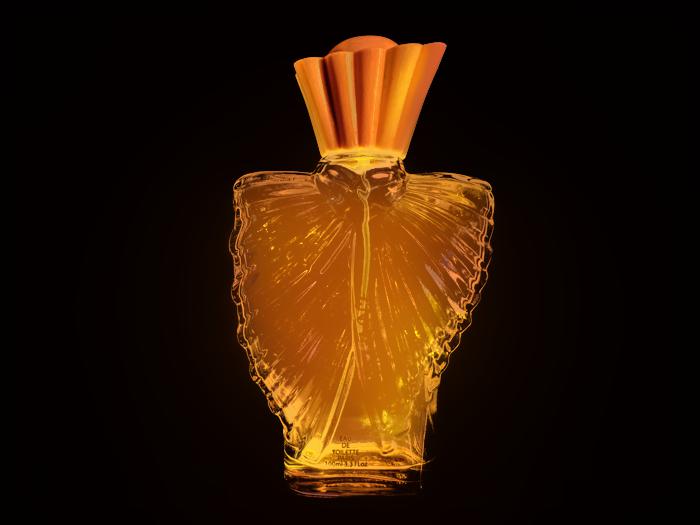} &
			\includegraphics[height = .13\linewidth, width = .16\linewidth]{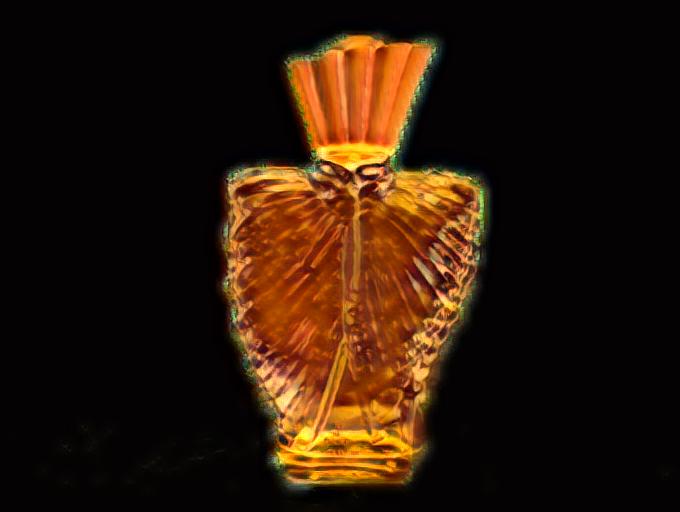} &
		\includegraphics[height = .13\linewidth, width = .16\linewidth]{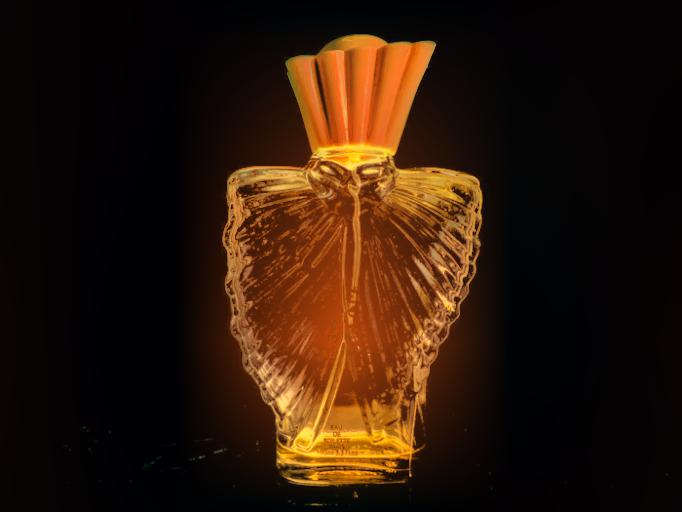}  \\
		
		\includegraphics[height = .1\linewidth, width = .16\linewidth]{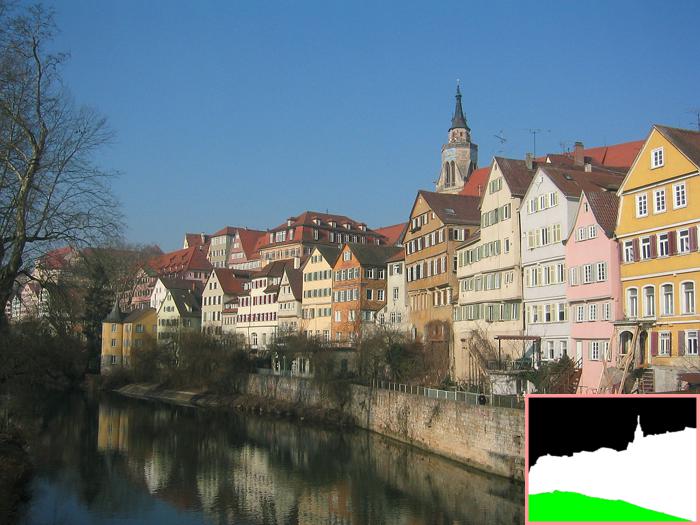} &
		\includegraphics[height = .1\linewidth, width = .16\linewidth]{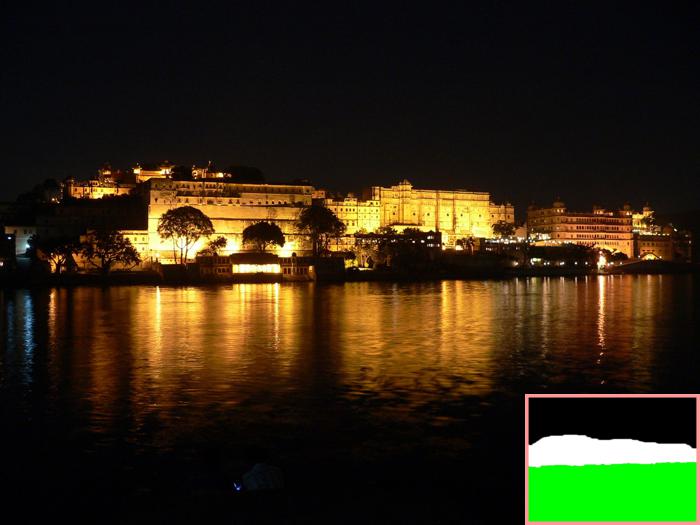} &
	    \includegraphics[height = .1\linewidth, width = .16\linewidth]{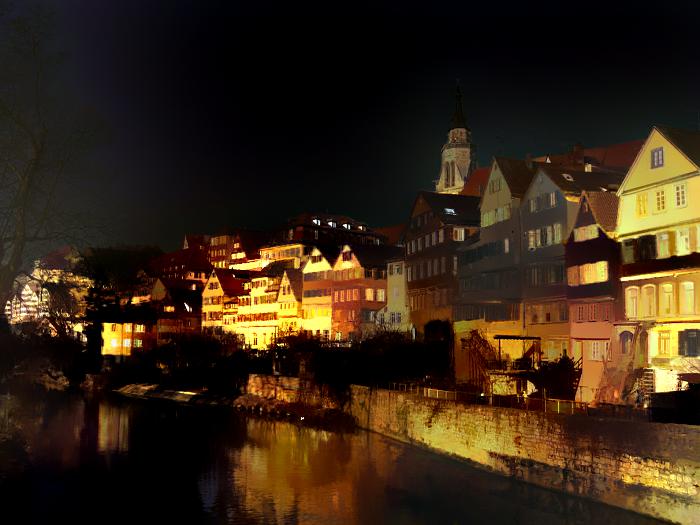} &
		\includegraphics[height = .1\linewidth, width = .16\linewidth]{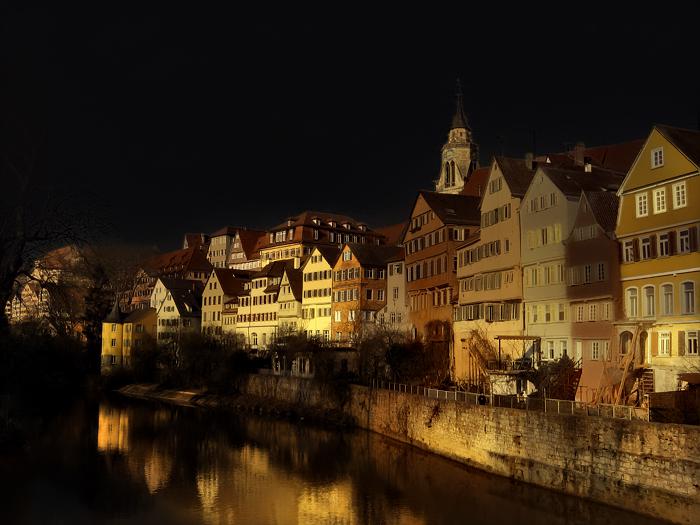} &
		\includegraphics[height = .1\linewidth, width = .16\linewidth]{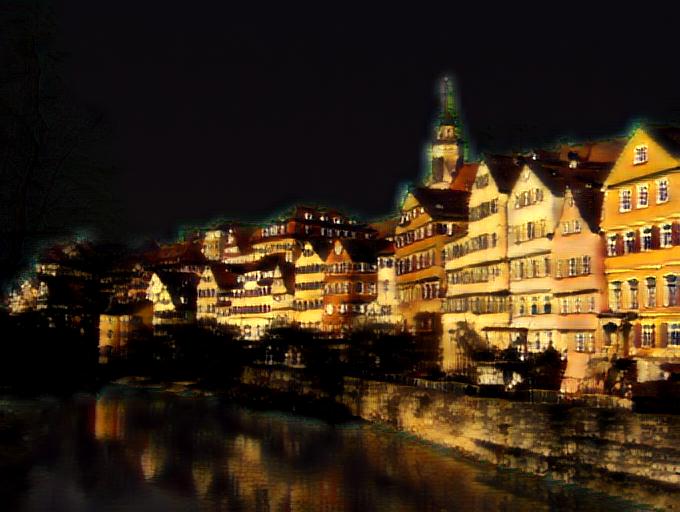} &
		\includegraphics[height = .1\linewidth, width = .16\linewidth]{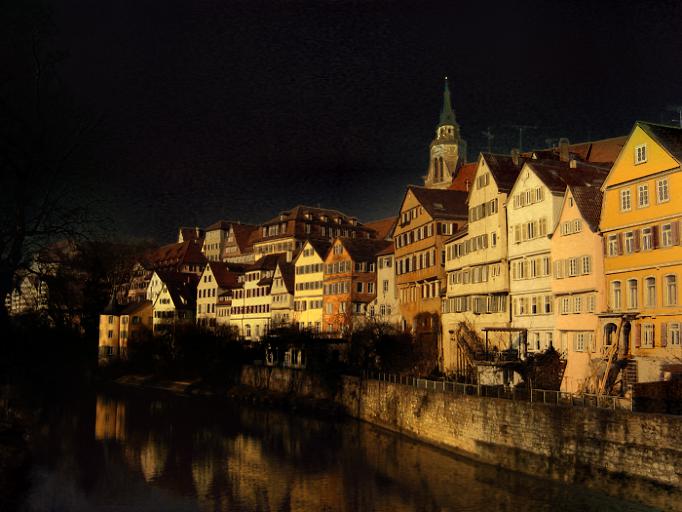}  \\
		
		{\footnotesize Content }& {\footnotesize Style} & {\footnotesize Luan~\cite{luan2017deep} }& {\footnotesize Li~\cite{li2018photoWCT} }& {\footnotesize Ours } & {\footnotesize +Bilateral } \\
	\end{tabular}
	\label{fig:photo_compare}
\end{figure*}

\vspace{1mm}
\noindent \textbf{Photo-realistic style transfer.}
Our algorithm can also produce un-distorted image style transfer results, due to the affinity preservation, which is particularly useful for photo-realistic style transfer.
We transfer the corresponding regions w.r.t a given mask for every content/style pairs.
This is done by separating the masked regions only within the transformation module (see Figure~\ref{fig:overview}), and then combining the transformed features of different regions, according to the mask.
We show the qualitative comparison with several recent work~\cite{luan2017deep,li2018photoWCT} in Figure~\ref{fig:photo_compare}.
When applying an auto-encoder with $relu3\_1$ as the bottleneck, the model is able to generate un-distorted stylized images (see column 5 in Figure~\ref{fig:photo_compare}), which have comparable visual qualities to those methods that are particularly targeting on photo-realistic stylization~\cite{luan2017deep,li2018photoWCT}. 
Note that both~\cite{luan2017deep} (186 seconds for 512$\times$256 images) and \cite{li2018photoWCT} (2.95 seconds for 512$\times$256 images) are more than 2 orders of magnitude slower than the proposed method (0.025 seconds for 512$\times$256 images), due to the time-consuming propagation modules that are designed to explicitly preserve the image affinity.
Our results can be further processed by a simple filter such as a joint bilateral filter or bilateral guided upsampling~\cite{chen2016bilateral} method to render final photo-realistic results with better object boundaries (see Figure~\ref{fig:photo_compare}).

\begin{figure*}[h]
	\centering
	\begin{tabular}{c@{\hspace{0.005\linewidth}}c@{\hspace{0.005\linewidth}}c@{\hspace{0.005\linewidth}}c@{\hspace{0.005\linewidth}}c@{\hspace{0.005\linewidth}}c}
		
		\includegraphics[height = .106\linewidth, width = .192\linewidth]{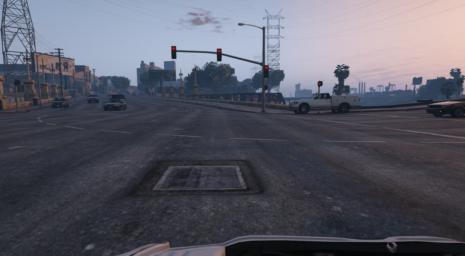} &
		\includegraphics[height = .106\linewidth, width = .192\linewidth]{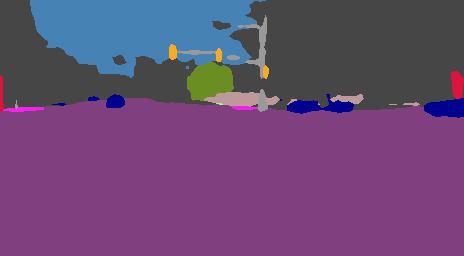} &
		\includegraphics[height = .106\linewidth, width = .192\linewidth]{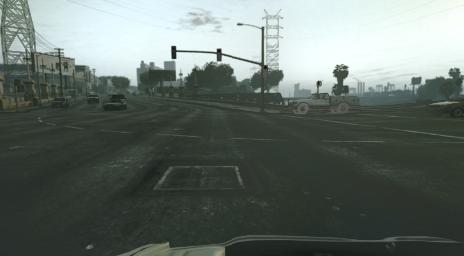} &
		\includegraphics[height = .106\linewidth, width = .192\linewidth]{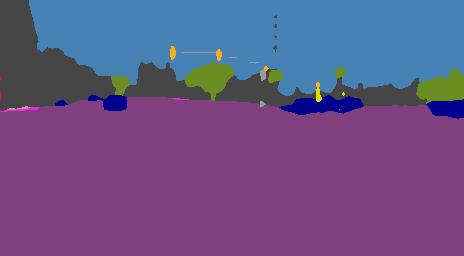} &
		\includegraphics[height = .106\linewidth, width = .192\linewidth]{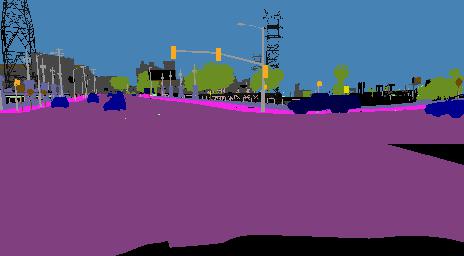} & \\
		
		\includegraphics[height = .106\linewidth, width = .192\linewidth]{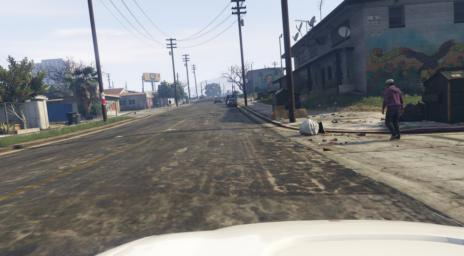} &
		\includegraphics[height = .106\linewidth, width = .192\linewidth]{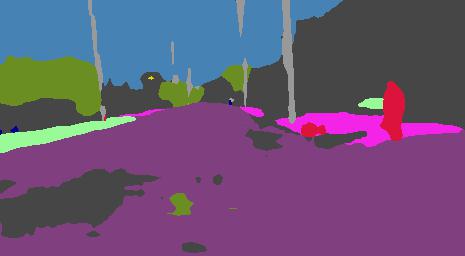} &
		\includegraphics[height = .106\linewidth, width = .192\linewidth]{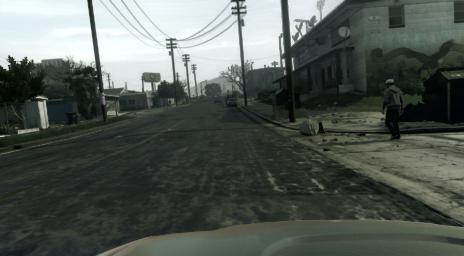} &
		\includegraphics[height = .106\linewidth, width = .192\linewidth]{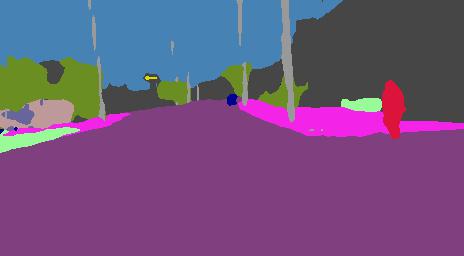} &
		\includegraphics[height = .106\linewidth, width = .192\linewidth]{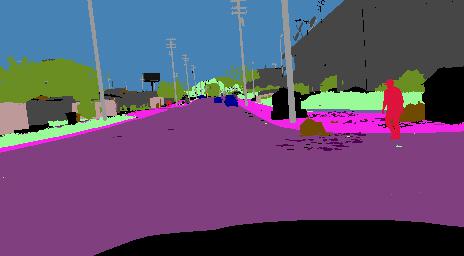} & \\
		
		\includegraphics[height = .106\linewidth, width = .192\linewidth]{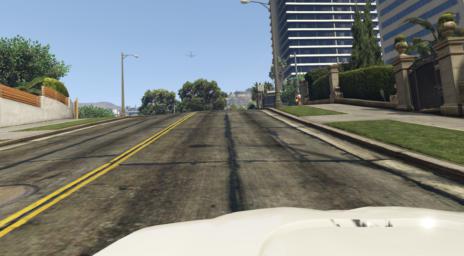} &
		\includegraphics[height = .106\linewidth, width = .192\linewidth]{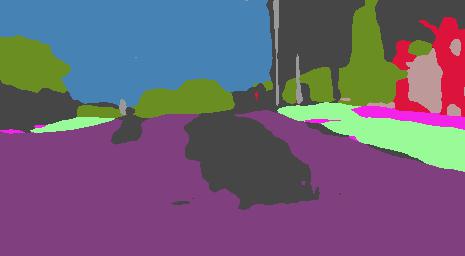} &
		\includegraphics[height = .106\linewidth, width = .192\linewidth]{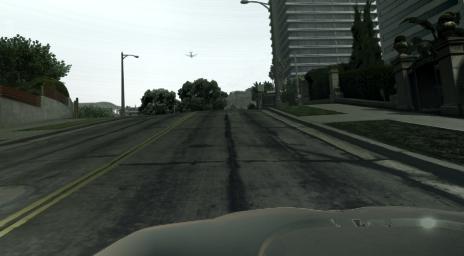} &
		\includegraphics[height = .106\linewidth, width = .192\linewidth]{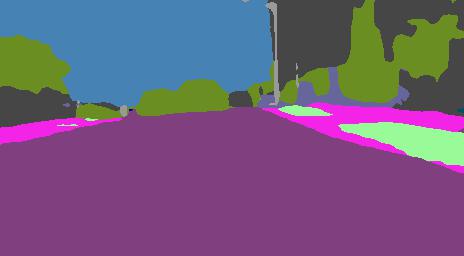} &
		\includegraphics[height = .106\linewidth, width = .192\linewidth]{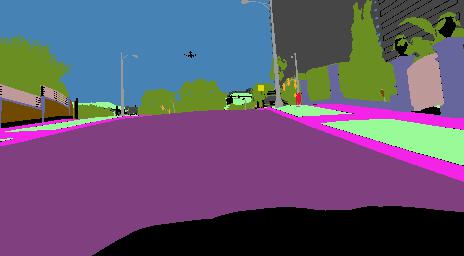} & \\
		
		{ Game} & { Game seg}& {  Transfer}& { Transfer  seg } & {  Ground truth } &\\
	\end{tabular}
		\vspace{-3mm}
	\caption{\footnotesize{Segmentation results by the PSPNet~\cite{zhao2017pspnet} before and after domain adaptation by our method. 
	Column 1 shows game images while column 3 shows transferred results by taking images from column 1 as content inputs and images from the Cityscapes~\cite{Cordts2016Cityscapes} dataset as style inputs.
	The segmentation results in column 4 are better than results in column 2, showing the usefulness of our algorithm in domain adaptation.}}
	\label{fig:domain_adap}
\end{figure*}

\subsection{Domain Adaptation: from Game to Real.}
\vspace{-1mm}
The linear transformation based arbitrary style transfer shares a lot of resemblance with typical high-level ideas of domain adaptation (e.g., by Sun et al.~\cite{coral}), in which both are proposed to align the second-order statistics between two different domains, e.g., style and content images~\cite{li_mmd}.
A pioneer work proposed by Liu et.al~\cite{domainadp2018} successfully utilized the WCT-based style transfer technique in the game-to-real domain adaptation with an unsupervised setting.
In this section, we validate that the proposed model can significantly narrow down the domain gap for semantic segmentation between the game images and the real images.

First, we apply the PSPNet semantic segmentation model~\cite{zhao2017pspnet}, which is well-trained on the Cityscapes~\cite{Cordts2016Cityscapes} dataset, to segment the images from the GTA~\cite{Richter_2016_ECCV} which contains similar traffic scenes, but with different image quality of computer games.
We then transfer the game images to ``real'' images, based on the proposed method.
Specifically, we implement the task as photo-realistic style transfer by randomly taking one image from the GTA dataset as content input, and another image from the Cityscapes dataset as style inputs.
The transformation is conducted w.r.t the corresponding semantic mask, where all non-shared labels from both image are considered as belonging to an additional class.
We then process semantic segmentation using the same PSPNet model on the transferred results, as shown in Figure~\ref{fig:domain_adap}.
%
The segmentation results on column 2 (segmentation results before domain adaptation) and column 4 (segmentation results after domain adaptation) demonstrate that the proposed algorithm is effective to adapt the domain of game images to real images in the semantic space.
We note that narrowing the gaps between these two datasets (especially adapting games images to real world scenes) is practical for increasing training samples for semantic/instance-level segmentation, flow and depth estimation, to name a few, due to the fact that the dense annotations of these tasks in the real-world scenes are difficult to obtain. 
Our algorithm provide an efficient and low-cost solution to generate numerous training data,  which could potentially benefit for various vision problems.

\section{Conclusions}
\vspace{-2mm}
In this work, we develop a theoretical foundation for analyzing arbitrary style transfer frameworks and present an effective and efficient algorithm by learning linear transformations. 
Our algorithm is computationally efficient, flexible for numerous tasks, and effective for stylizing images and videos. 
Experimental results demonstrate that the proposed algorithm performs favorably against the state-of-the-art methods on style transfer of images and videos.


\bibliographystyle{plain}

\end{document}